\documentclass{IEEEtran}

\usepackage[final]{graphicx}
\usepackage[cmex10]{amsmath}
\usepackage{amssymb}
\usepackage{graphics} 
\usepackage{epsfig}
\usepackage[tight,footnotesize]{subfigure}

\begin{document}
\title{Neuro-Fuzzy Computing System with the Capacity of Implementation on Memristor-Crossbar and Optimization-Free Hardware Training}

\author{Farnood~Merrikh-Bayat,
        Farshad~Merrikh-Bayat,
        and~Saeed~Bagheri~Shouraki
\thanks{Farnood Merrikh-Bayat and Saeed Bagheri Shouraki are with the Department
of Electrical Engineering, Sharif University of Technology, Tehran, Iran, e-mails: f\_ merrikhbayat@ee.sharif.edu, bagheri-s@sharif.edu.}
\thanks{Farshad Merrikh-Bayat is with the Department
of Electrical and Computer Engineering, University of Zanjan, Zanjan, Iran, e-mail: f.bayat@znu.ac.ir.}}
 \maketitle

\begin{abstract}
In this paper, first we present a new explanation for the relation
between logical circuits and artificial neural networks, logical
circuits and fuzzy logic, and artificial neural networks and fuzzy
inference systems. Then, based on these results, we propose a new
neuro-fuzzy computing system which can effectively be implemented
on the memristor-crossbar structure. One important feature of the
proposed system is that its hardware can directly be trained using
the Hebbian learning rule and without the need to any
optimization. The system also has a very good capability to deal
with huge number of input-out training data without facing
problems like overtraining.
\end{abstract}

\maketitle


\section{Introduction}

During past years, lots of efforts have been made to approach to
the computing power of human brain. These efforts roughly can be
categorized into several different areas such as Artificial Neural
Networks (ANNs), Fuzzy Logic, \textit{etc.} By reviewing these
works we can simply recognize that most of them have only
concentrated on the software and we can rarely find a good sample
for hardware implementation of an intelligent system. In addition,
lots of the suggested structures or methods do not have biological
support. By considering the number of neurons in the human brain
and the complexity of connections between them, the importance of
having an efficient hardware with the ability of expanding into
that scale becomes more and more clear. According to the nature of
computation and memory in brain, now it is a well-accepted fact
that this hardware should be in analog form since in this case it
can work much faster than conventional digital circuits. However,
heretofore, there was a big obstacle in front of reaching this
goal. In fact, there was no simple passive element that can be
used for storing and manipulation of data like synaptic weights.
Note that although analog values can be stored in capacitors as
voltage or charge, the stored values cannot easily be read and
used in computations without being altered. In addition, according
to the leakage problem the data stored in capacitors will vary in
time. As a result, most of the analog hardwares proposed so far
are somehow inefficient and area consuming designs (see, for
example \cite{Botros,Dinu,Kuo}). 

In 2008 and after the physical realization of first \textit{memristive device} or \textit{memristor} \cite{williams} (used interchangeably in the rest of paper) which was also predicted by Leon Chua in 1971 \cite{Chua}, the field of brain emulation has been revived.  This is mainly because of the existence of some similarities between the physical behavior of memristor and synapses in brain \cite{Snider,Jo}. Memristor is a passive device whose properties such as resistance (known as memristance) or conductance (known as memductance) can be changed by applying a suitable voltage or current to it. Therefore, analog values such as synaptic weights can be stored in this device by tuning its memristance. Fortunately, unlike capacitor, memristor can retain its memristance for a long period of time in the absence of the voltage or current applied to it \cite{Chang2}. Moreover, since memristor simply acts  as a time
varying resistor it can easily be applied in many classical
circuit designs.

Aforementioned properties of memristor motivated some researchers to develop new brain-like computing architectures and methods where the main focus was on the simplicity of the resulted memristor-based hardware. The conducted studies on this subject can be divided into two main categories. The first category belongs to the works trying to implement fuzzy inference methods by using memristive hardwares \cite{Farnood1,Farnood2,versachi}. For example, in \cite{Farnood2} we showed that fuzzy relations (describing the relation between input and output fuzzy concepts or variables in an imprecise manner) can efficiently be  formed on memristor crossbar structures by using Hebbian learning method.  Moreover, we also showed that the system constructed by  concatenation of these basic units can perform fuzzy computations. The second category consists of the studies concentrated on  hardware implementation of spiking neural networks and their learning methods such as Spike Timing-Dependent Plasticity (STDP) \cite{Gerstner} using memristor crossbar structures \cite{Kim,Cantley,Afifi,Zamarre,Adhikari}. In spite of their popularity, these networks suffer from some disadvantages. Firstly, there is no guarantee or proof for the convergence of the methods used for their training. Secondly, no perfect method has been proposed so far for training multi-layer spiking neural
networks. This is why we do not see applications like function approximation (which requires multi-layer networks)  to be implemented by these networks, although they exhibit excellent
performance in some other applications like data classification \cite{Zamarre}. Thirdly, spiking neurons have parameters (like threshold value of neurons) to be set. Finally, in these networks connection weights can be either positive or negative, which is a disadvantage from the hardware-implementation point  of view.

In this paper, we propose a new computing system which
leads to a simple hardware implementation and can remove some of
the aforementioned difficulties. For this purpose, we start from logical circuits as the simplest multi-layer networks and reveal some of the similarities and dissimilarities between them and ANNs. For example, it will be demonstrated that logical circuits
can be considered as networks whose connections can partly be
tuned by using Hebbian learning rule \cite{Hebb}. This provides us with some ideas for improvement of the performance of ANNs. Then, we show that even without changing the structure of conventional ANNs, their working procedure can be explained based on fuzzy concepts. It means that \textit{fuzzy inference systems  have biological support}. Combination of these findings leads us to a new multi-layer neuro-fuzzy computing system with very interesting and important properties as summarized below. First of all, the proposed system can be trained without using any optimization methods. Second, it accepts inputs in fuzzy format and generates fuzzy outputs. Third, all connection weights in the proposed method are non-negative. Fourth, neurons of the network do not have any parameter to be tuned. Fifth, in the proposed structure computing and memory units are assimilated with each other like what we see in human brain. Finally, it will be shown in the rest of the paper, the most important advantage of our method is that it can be simply implemented by using memristor crossbar structures. It is worth to mention that, roughly speaking,
complexity of the hardware needed to implement the proposed
neuro-fuzzy computing system is almost the same as the hardware
needed to implement a typical spiking neural network.

The rest of this paper is organized as follows.  In Section \ref{Rela} we review some similarities and dissimilarities between logical circuits and neural networks to gain some insights about how we can improve learning algorithms of neural networks. Then, by extending digital logic in a way that it can work with continuous variables we reach to fuzzy logic. Finally, in the same section we show that the working procedure of neural networks can be explained based on fuzzy concepts which means that fuzzy inference systems can have biological support like neural networks. In Section \ref{ourmethod} we develop our own neuro-fuzzy computing system and its associated learning method. Hardware implementation of the proposed method based on memristor crossbar structures is described in Section \ref{hardwa}. Section \ref{simres} is devoted to the presentation of simulation results before conclusion in Section \ref{conclu}.

\section{The relations between digital logic, fuzzy logic and neural network}\label{Rela}

\subsection{Similarities and dissimilarities between logical
circuits and artificial neural networks}\label{logics}

\begin{figure}[!t]
\centering 
{
\includegraphics[width=3.2in,height=2.2in]{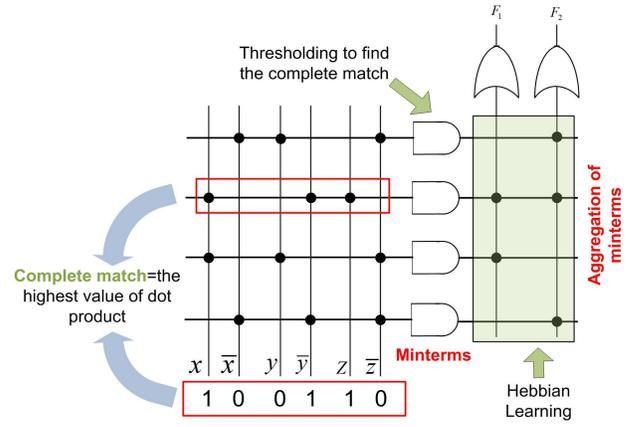}}
\caption{Typical logical circuit implementing two logical function in a standard form of sum of products.}
\label{fig1} 
\end{figure}

Logical circuits can be considered as the simplest form of
multi-layer networks. To show this, first note that any binary
function can be written in the standard form of sum of products
(min-terms) \cite{Nelson}. For this purpose, one can simply \emph{add} (more
precisely, OR) those min-terms that activate the function under
consideration. For example, Figure \ref{fig1} shows the structure used
to construct the sample binary functions $F_1$ and $F_2$. In this
figure, $x$, $y$, and $z$ are the binary input variables and each
logical gate acts only on the signals entered to it from the
cross-points denoted by black circles.

The simple logical structure shown in Fig. \ref{fig1} acts very similar to the
conventional ANNs. In fact, such a structure can be considered as
a two-layer network. The first part of this structure, which
consists of input and hidden layers (\textit{i.e.}, the layer consists of
AND gates) is used to create the min-terms and the second part,
which consists of hidden and output layers (\textit{i.e.}, the layer
consists of OR gates) is used to add the products and generate the
final outputs. It is well-known that any binary function can be
constructed by using such a three-layer network. Clearly, similar to conventional ANNs \cite{Fausett}, in the
structure shown in Fig. \ref{fig1} increasing the number of layers will not
enhance the accuracy of the resulted binary function.

In order to gain more insight about the similarities between ANNs and logical circuits, we can describe the role of the first two layers of the circuit
shown in Fig. \ref{fig1} in another manner. As it can be observed, each
input of this circuit is multiplied in a specific weight, which is
equal to either 1 or 0, and then it is entered to the hidden
layer. Denote the weight matrix multiplied to the variables of
input layer to form the variables of hidden layer as
$\mathbf{W}_1^\mathrm{logic}$. Clearly, each row of this matrix,
which is stored on cross-points of the first crossbar, corresponds
to a certain combination of input variables. More precisely, the
rows of $\mathbf{W}_1^\mathrm{logic}$ determine the min-terms used
to create the functions at the system output, which are independent from the definition of output functions. Since $n$ input
variables can generate, in general, $2^n$ min-terms, number of the
rows of $\mathbf{W}_1^\mathrm{logic}$ is always smaller than or equal to
$2^n$. However, as a general
observation, digital functions usually do not need all of these
$2^n$ min-terms to be constructed.

According to the above discussions, it seems that when the circuit
shown in Fig. \ref{fig1} is subjected to a certain binary input string, 
internal product of input variables and the pattern stored at each
row of the crossbar (which corresponds to a specific min-term) is
performed to determine the similarity between these two binary
strings. Then each AND gate acts somehow as a neuron with hard-thresholding
activation function (in contrast to OR gates which are soft-thresholding operators): Only the output of the AND gate (neuron) located on the row with
complete agreement with input variables is set to 1 and the output
of other AND gates (neurons) is set to 0. Note that although each AND gate
here acts as a thresholding operator, unlike the activating
function of neurons in ANNs, its thresholding level is determined
beforehand and cannot be changed in practice. For example, when we have two input variables and one 3-input AND gate, we should set one of its input terminals equal to logic 1 to make sure that the activation of other two  inputs can pass the threshold of the gate. Here, it is worth to mention that by
recognizing the AND gate (or equivalently the min-term) with activated output in Fig. \ref{fig1} one can
determine the activated inputs and consequently, specify the
concepts and events \textit{occurred simultaneously}. In other words, it can be said that the activation of each min-term indicates the simultaneous occurrence of certain concepts (represented by input terminals). As it will be shown
later, this method of determining the simultaneous happening of
different events is of essential importance in designing our
proposed neuro-fuzzy computing system.

In the second part of the circuit shown in Fig. \ref{fig1} the min-terms
generated in the first part are combined together to produce the
final outputs of the system. But, unlike the first part, the
weights connecting the AND gates of hidden layer to the OR gates
of output layer are not predictable and fully dependent on the
definition of binary functions. However, interesting point is that
these weights can be adjusted following the \textit{Hebbian learning rule} \cite{Hebb}
used in ANNs. More precisely, it can be easily verified that when
for a certain input the output of system is equal to logic 1,
output of the activated AND gate (neuron) is connected to the activated
output. It means that by the simultaneous applying the input and
output data to input and output layers respectively, and using the
Hebbian learning rule, the connections of the second part of the logical
circuit can be constructed automatically. In addition, note that similar to most of the ANNs, thresholding strength of output neurons (\textit{i.e.} OR gates) is weaker than the thresholding strength of neurons of the hidden layer (\textit{i.e.} AND gates).

Although multi-layer logical circuits act very similar to ANNs,
they have considerable advantages that we are trying to use them in our proposed computing system. Firstly, in logical circuits
all input and output signals, as well as the weights stored in
cross-points, are non-negative in nature. Secondly and more importantly, unlike ANNs,
any binary function can be realized by training only the weights
stored in the second crossbar of the two-layer logical circuit
shown in Fig. \ref{fig1} without using any optimization methods (recall that the pattern stored in the first
crossbar is independent of the special binary functions aimed to
be constructed, and moreover, this pattern can be determined only
by knowing the number of input variables). Thirdly, assuming that
$m_i$ and $\vee$ stand for the $i$th min-term and the OR
operator respectively, we have $m_i\vee m_i=m_i$. It yields
that repeated min-terms (\textit{i.e.} putting emphasis on the simultaneous occurrence of several concepts or inputs) in the definition of a binary function
(which is expressed in the form of sum of products) has no effect
on the input-output relation of the resulted structure. In other
words, assuming that the weights stored in the second crossbar are
adjusted by using the Hebbian learning method, subjecting the
system to repeated input-output data will not change anything in
the system. Clearly, it is an important capability that ANNs do
not have. Fourthly, any binary function can be expressed in the
form of sum of products. This provides us with an easy way to explain and
understand the role of binary functions based on the concepts  occurring simultaneously since it is similar to the
logic used by the brain of humankind. In other words, one can
easily discover the task of a binary function expressed in the
form of sum of products, while it is quite difficult to get such a
knowledge by decoding the synaptic weights of the given ANN.

Finally, it should be noted that although using the complement of
a binary variable provides us with no more information, it is
still a common practice to use both a binary variable and its
complement in logical circuit design. In fact, it seems that in
the logical circuit design any distinct value of the input
variable is considered as an individual or independent concept, which is probably
because of the fact that our brain prefers to work with two
contradicting concepts than a non-contradicting one. This is completely in contrast with ANNs in which we usually use only one input for each continuous input variable. In other words, similar to fuzzy logic \cite{Farnood2}, here it seems that we are considering one input terminal per each distinct value of input variable and the input that we apply to any of these terminals somehow is the representative of our confidence degree about the occurrence of its corresponding concept.   For this reason, in the following discussions one input terminal is considered for each distinct value the input or output variable can take.

\subsection{The relation between logical circuits and fuzzy
logic}\label{sec_fuzzy} One main question that arises at this point is:
``Can we extend the multi-layer logical circuit shown in Fig. \ref{fig1}
such that it can work with non-binary input variables?".
Fortunately, the answer of this question is positive thanks to the
fuzzy logic. In fact, the method used to express a binary function
in the form of sum of products is very similar to the method of
constructing a continuous function based on fuzzy rules. In the following we discuss on
this subject with more details.

First, consider a multi-input-single-output fuzzy inference system
whose input-output relation can be considered as the mathematical
map $\mathbf{X}\rightarrow Y$, where
$\mathbf{X}\subset\mathbb{R}^n$ and $Y\subset\mathbb{R}$.
Moreover, assume that the output of this fuzzy inference system is
obtained by aggregation of the output of $N$ fuzzy rules in the
form of \cite{leszek}:
\begin{eqnarray}\label{rule_k}
R^{(k)}: \mathrm{IF}~ x_1~ \mathrm{is}~ A_1^k~ \mathrm{AND}~ x_2~
\mathrm{is}~ A_2^k ~\mathrm{AND}~ \ldots~ x_n ~\mathrm{is}~ A_n^k~\nonumber\\
\qquad \qquad \mathrm{THEN}~ y ~\mathrm{is}~ B^k,\quad k=1,\ldots,N,
\end{eqnarray}
which, for better explanation of the similarities between fuzzy
logic and logical circuits, can be rewritten as:
\begin{equation}\label{fuz_rule}
R^{(k)}: \mathrm{IF}~ \mathbf{x}~ \mathrm{is}~
\mathbf{\Sigma}_n^k~ \mathrm{THEN}~ y ~\mathrm{is}~ B^k.
\end{equation}
where $x_i$ ($i=1,2,\ldots,n$) and $y$ are input and output
variables, respectively, $\mathbf{x}=
[x_1,x_2,\ldots,x_n]\in\mathbf{X}$, $y\in Y$, $\{A_1^k,
A_2^k,\ldots, A_n^k\}$ are fuzzy sets with membership functions
$\mu_{A_i^k}(x_i)$ ($i=1,2,\ldots,n$; $k=1,2,\ldots,N$) defined on
the universal set of input variables, $B^k$ ($k=1,2,\ldots,N$) are
fuzzy sets defined on the universal set of the output variable,
and $\mathbf{\Sigma}_n^k$ is a n-dimensional fuzzy set with the following
multi-input membership function:
\begin{equation}\label{t_norm1}
\mu_{\mathbf{\Sigma}_n^k}(x_1,x_2,\ldots,x_n)=t\left(\mu_{A_1^k}(x_1)
,
\mu_{A_2^k}(x_2), \ldots, \mu_{A_n^k}(x_n)\right),
\end{equation}
where $t$ is a $t$-norm operator. Since the fuzzy propositions in
the antecedent part of (\ref{rule_k}) are combined using the fuzzy
AND operator, we call such rules \emph{AND-type} fuzzy rules. According to
the properties of the $t$-norm operator, the fuzzy proposition in
the consequent part of an AND-type fuzzy rule is true (in fuzzy
sense) if all of the fuzzy propositions in the antecedent part of
that rule are true (again in fuzzy sense). Similarly, combination
of the fuzzy propositions of the antecedent part with fuzzy OR leads
to an \emph{OR-type} fuzzy rule. Similar to logical circuits,
AND-type fuzzy rules are often preferred since our brains find
them more reasonable. Here, we can see the structural and behavioral similarities between fuzzy logic and logical circuits. For example, antecedent part of (\ref{rule_k}) plays the role of the creation of min-terms in the first part of the circuit shown in Fig. \ref{fig1} but with the difference that here $A^k_i$s are fuzzy numbers (note that in logical circuits $A^k_i$s were crisp numbers chosen from the set $\{0, 1\}$). In addition, aggregation of the output of $N$ fuzzy rules by using a $s$-norm operator is similar to the function of the second part of the logical circuit shown in Fig. \ref{fig1}. Finally, note that in fuzzy logic all inputs, outputs and coefficients are non-negative as well.

Now, let us look at the working procedure of each fuzzy rule in another way. According to (\ref{fuz_rule}) and (\ref{t_norm1}) we interpret the antecedent part of each fuzzy rule
as a subspace of $\mathbb{R}^n$ (denoted as $\mathbf{\Sigma}_n^k$) with non-crisp (fuzzy)
borders. In fact, $\mathbf{x}^\ast\in\mathbb{R}^n$ \emph{belongs}
to the region defined by antecedent part of (\ref{fuz_rule}) if
$\mu_{\mathbf{\Sigma}_n^k}(\mathbf{x}^\ast)$ is a \emph{big}
number (\textit{i.e.}, it is \emph{close} to unity), and vice versa. More precisely, it can be said that any $\mathbf{x}^\ast\in\mathbb{R}^n$ belongs to the subspace $\mathbf{\Sigma}_n^k$ with the confidence degree of $\mu_{\mathbf{\Sigma}_n^k}(\mathbf{x}^\ast)$ (for $k=1,2, \ldots, N$).
Clearly, lying the point corresponds to a crisp input data in
any of these subspaces indicates the simultaneous happening of
certain fuzzy concepts defined by these subspaces (or equivalently, the corresponding $A^k_n$s). For this reason, we call the subspace
specified by antecedent part of each fuzzy rule a \emph{fuzzy
min-term} (in the special case when $A_i^k$ is a fuzzy set with
singleton membership function from the support set of $\{0,1\}$, each fuzzy min-term is reduced to a logical min-term denoted as a
single point in the space of input variables). Considering the fact that
each of the fuzzy sets used in antecedent part of fuzzy rules can
have a different membership function and support set, infinite
number of unique fuzzy min-terms can be defined on the
$n$-dimensional space of input variables. But, fortunately, in
most of the real-world applications the input data are accumulated
in certain parts of the space of input variables, and
consequently, it is sufficient to define the limited number of fuzzy min-terms such
that they cover only those areas. In other words, although the
number of fuzzy min-terms theoretically can be very large,
commonly only a few number of these min-terms have a considerable
influence on the system output. Hence, in order to construct a
fuzzy inference system we need to identify only the most important
fuzzy min-terms (or fuzzy rules).


Now consider the case in which the $k$th rule given in
(\ref{rule_k}) is subjected to the following fuzzy input data:
\begin{equation}\label{obs_data1}
q:~ x_1~ \mathrm{is}~ A'_1~ \mathrm{AND}~ x_2~ \mathrm{is}~ A'_2
~\mathrm{AND}~ \ldots \mathrm{AND}~x_n~ \mathrm{is}~ A'_n,
\end{equation}
which can equivalently be expressed as:
\begin{equation}\label{obs_data2}
q:~\mathbf{x}~\mathrm{is}~\mathbf{\Sigma}_n',
\end{equation}
where $\{A_1',A_2',\ldots,A_n'\}$ are fuzzy sets with membership
functions $\mu_{A_i'}(x_i)$ ($i=1,2,\ldots,n$), and
$\mathbf{\Sigma}_n'$ is a fuzzy set with multi-variable membership
function $\mu_{\mathbf{\Sigma}_n'}(\mathbf{x})$. The input data
given in (\ref{obs_data1}) stimulates the antecedent part of each
of the fuzzy rules given in (\ref{rule_k}) to a certain degree.
Clearly, the amount of activation of each fuzzy rule determines
the contribution of the consequent part of that rule in the final
output. Various methods are available to determine the amount of
activation of the antecedent part of a fuzzy rule for the given
fuzzy input data \cite{leszek}. However, in the following we want to deal
with this issue from a different point of view.

Consider again the fuzzy input data given in (\ref{obs_data2})  which specifies the subspace $\mathbf{\Sigma}_n'$ with fuzzy borders. In
general, this  subspace overlaps with
the subspace of each of the fuzzy min-terms, \textit{i.e.} $\mathbf{\Sigma}_n^k$s, in the
$n$-dimensional space to a certain degree. But, unlike the binary
logic, $\mathbf{\Sigma}_n'$ most likely does not completely
overlap with any of the $\mathbf{\Sigma}_n^k$s ($k=1,\ldots,
N$) and consequently, here it is reasonable to assume that various
fuzzy min-terms are activated to different degrees when the system
is subjected to this fuzzy input. More precisely, the degree of
activation of the $k$th fuzzy rule is proportional to the amount
of overlapping between subspaces $\mathbf{\Sigma}_n^k$ and
$\mathbf{\Sigma}_n'$. So, similar to the binary logic, the fuzzy
inference system first evaluates the similarity between the fuzzy
input data and fuzzy min-terms and then applies a kind of
soft-thresholding function (i.e., the
$s$-norm operator) to determine the contribution of the output
corresponding to each fuzzy min-term in the final outcome.

Here, we tried to show some of the similarities between fuzzy logic and logical circuits. On the other hand, in previous section we demonstrated how the aggregation of logical min-terms can be performed by using Hebbian learning rule. Therefore, it seems that by inspiration from logical circuits, we can propose a simple method to create fuzzy rules automatically based on  the available training data. However, the main importance of the proposed  method relates to its ability to work with ANNs. In fact, in the rest of this paper we will show that our proposed method can be used to train large scale ANNs, which can be considered as a big step toward the emulation of the computing power of human brain. This is mostly because of the fact that, as we will show in the next section, ANNs can be considered as systems that work with fuzzy concepts.


\subsection{Interpretation of ANNs as fuzzy inference systems}
The aim of this section is to provide an answer to the following
questions from a new point of view: (1) How similar is the
behavior of ANNs to fuzzy inference systems? (2) How can we
effectively and optimally determine and implement the fuzzy
min-terms and consequently, inference engine of a very large-scale system? By answering the former question we can show that fuzzy inference systems have biological support. In order to answer these questions and finally propose our method for hardware
implementation of neuro-fuzzy computing systems first we 
describe the function of ANNs based on fuzzy concepts in a new and interesting  manner.

\begin{figure}[!t]
\centering 
{
\includegraphics[width=3.2in,height=2.2in]{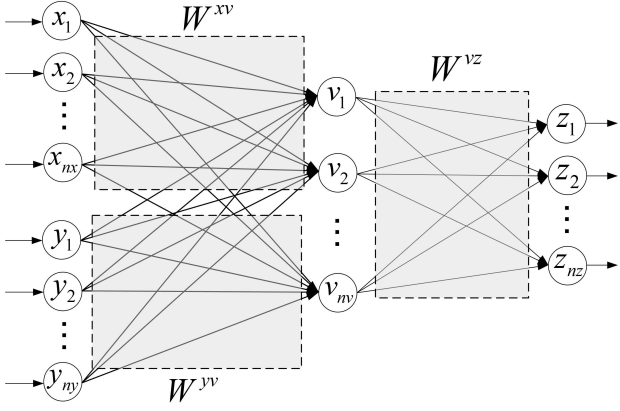}}
\caption{Typical artificial neural network with one hidden layer. In these systems, connection weights are usually determined during the learning process by the use of optimization methods.}
\label{fig2} 
\end{figure}

Consider the two-layer ANN shown in Fig. \ref{fig2}. In this figure:
\begin{itemize}
    \item $\mathbf{x}=[x_1,x_2,\ldots,x_{nx}]$ and
    $\mathbf{y}=[y_1,y_2,\ldots,y_{ny}]$ are the vector
    of input variables entered to the input layer,
    \item $\mathbf{v}=[v_1,v_2,\ldots,v_{nv}]$, where $v_i$ is the
    output of neuron number $i$ of hidden layer,
    \item $\mathbf{W}^{xv}=[w_{ij}^{xv}]_{nv\times nx}$ is the matrix
    containing the weights connecting the neurons of input layer to the neurons of hidden
    layer, where $w_{ij}^{xv}$ is the weight of synapse
    connecting the $j$th entry of $\mathbf{x}$, $x_j$, to the
    $i$th entry of $\mathbf{v}$, $v_i$. Using this
    notation, output of neuron $i$ of the hidden layer,
    $v_i$, is obtained as:
\begin{eqnarray}\label{ann1}
v_i&=&f\left(\sum_{l=1}^{nx}x_lw_{il}^{xv}+\sum_{l=1}^{ny}y_lw_{il}^{yv}\right)
\nonumber\\&=&f\left(\mathbf{w}_i^{xv}\mathbf{x}^T+\mathbf{w}_i^{yv}\mathbf{y}^T\right),~ 
i=1,2,\ldots,nv,
\end{eqnarray}
where
$\mathbf{w}_i^{xv}=[w_{i1}^{xv},w_{i2}^{xv},\ldots,w_{inx}^{xv}]$,
$\mathbf{w}_i^{yv}=[w_{i1}^{yv},w_{i2}^{yv},\ldots,w_{iny}^{yv}]$,
and $f(.)$ is the activation function of the neurons of hidden
layer.

    \item $\mathbf{W}^{vz}=[w_{ij}^{vz}]_{nz\times nv}$ is the matrix containing the weights
    connecting the neurons of hidden layer to the neurons of
    output layer, where $w_{ij}^{vz}$ is the weight of the synapse
    connecting the $j$th entry of $\mathbf{v}$, $v_j$, to the
    $i$th entry of $\mathbf{z}$, $z_i$.
    \item $z_i$ is the output of the $i$th neuron of output
    layer which can be expressed as follows:
    \begin{equation}
z_i=f\left(\sum_{l=1}^{nv}v_lw_{il}^{vz}\right)=f
\left({\mathbf{w}}_i^{vz}\mathbf{v}^T\right), \quad
i=1,2,\ldots, nz,
    \end{equation}
  where
    ${\mathbf{w}}_i^{vz}=[w_{i1}^{vz},w_{i2}^{vz},\ldots,w_{i,nv}^{vz}]$.
\end{itemize}

\begin{figure*}[!t]
\centering 
{
\includegraphics[width=6.5in,height=5in]{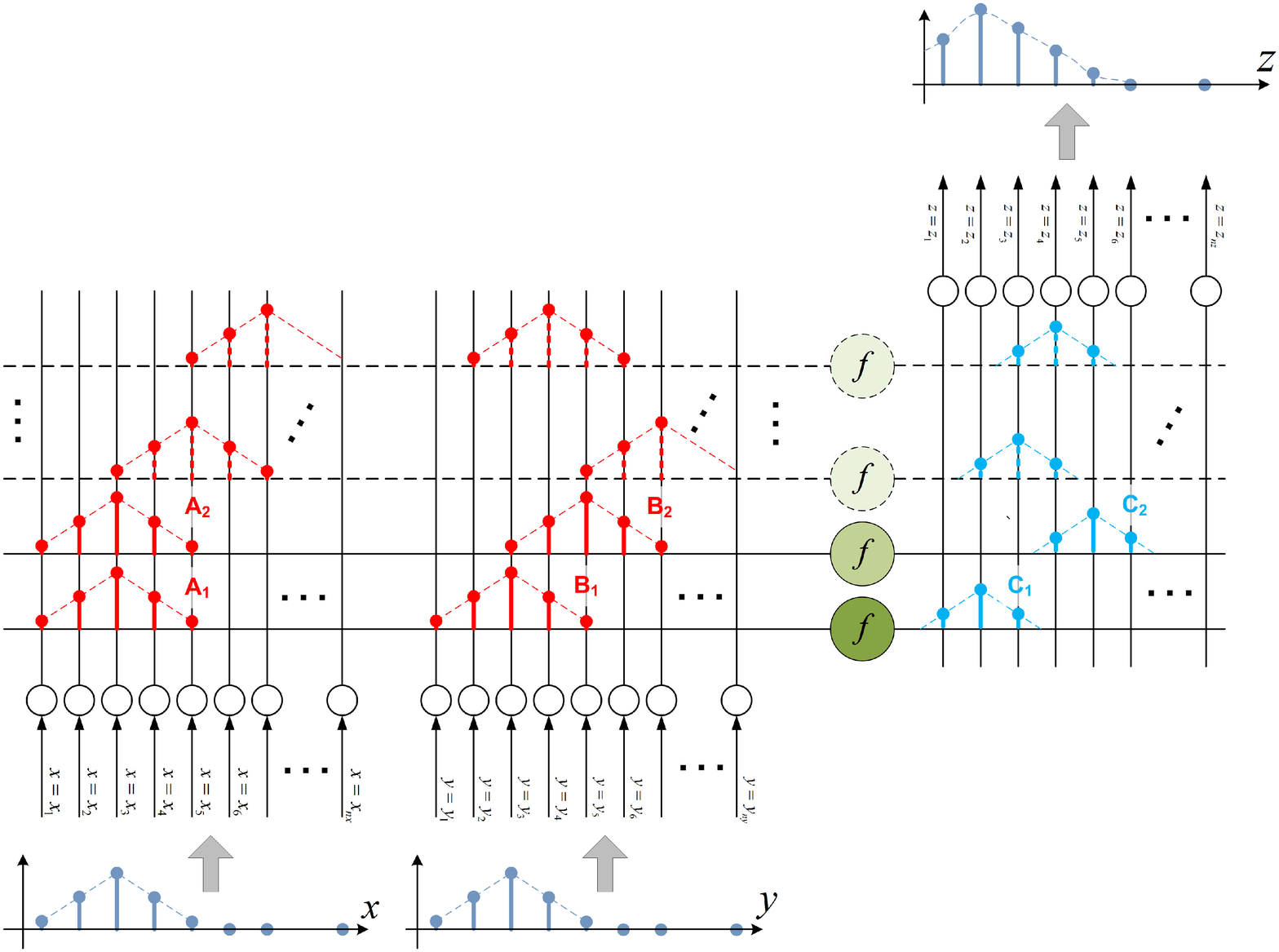}}
\caption{Simple realization of the ANN shown in Fig. \ref{fig2}  based on fuzzy concepts. In other words, this figure shows that it is possible to interpret the working procedure of conventional ANN similar to what we have in fuzzy inference system without changing its structure. In this
figure, each row of the structure implements a simple fuzzy rule or min-term. In fact, here we have assumed that connection weight matrices model the universal sets of input and output variables and by programming their entries correctly, any fuzzy set can be created on these universal sets.}
\label{fig3} 
\end{figure*}

Figure \ref{fig3} shows the realization of the ANN shown in Fig. \ref{fig2}. In this
figure, matrices of synaptic weights are implemented by using the
crossbar structure and it is assumed that the crossbars have the
property that generate the sum of products of input variables in
the weights stored at cross-points. For example, the signal
entered to each neuron of hidden layer is equal to the sum of the
product of signals at input layer to the synaptic weights stored
at cross-points (i.e., the signal entered
to the $i$th neuron of hidden layer is equal to
$\mathbf{w}_i^{xv}\mathbf{x}^T+\mathbf{w}_i^{yv}\mathbf{y}^T$). In this figure typical values are assigned to
synaptic weights and input data such that the height of each bar
is proportional to the value of the corresponding variable or
weight.

Interestingly, function of the structure shown in Fig. \ref{fig2} can be interpreted in
another way. In fact, similar to the case shown in Fig. \ref{fig1},
$x_1,\ldots,x_{nx}$ and their corresponding vertical wires in the first crossbar can be considered as the representative of the different values or concepts  the
linguistic input variable $x$ can take. We have a similar situation for variables $y$ and $z$. In this case, at the input and output terminals and at the rows of the structure we have constructed the discrete universal set of the variables $x$, $y$ and $z$.  Now, the special
values assigned to $x_1,\ldots,x_{nx}$ can be considered as the points of the
membership function of the fuzzy input proposition ``$x$ is $A'$". In other words, here it is assumed that the input applied to the $i$th input neuron shows our confidence degree about the occurrence of the concept assigned to that neuron for the observed  fuzzy input data.
Similarly, the synaptic weights stored at each row of the
crossbar located between input and hidden (hidden and output)
layer can be considered as the points of the membership function
of the antecedent (consequent) part of the corresponding fuzzy
rule. For example, Fig. \ref{fig3} shows implementation of the fuzzy sets
$A_1$ and $B_1$ (defined on the universal set of $x$ and $y$,
respectively) based on the coefficients stored at the lowest row
of the weight matrices $\mathbf{W}^{xv}$ and $\mathbf{W}^{yv}$,
respectively. Clearly, in this structure any new fuzzy rule can be
added to the fuzzy rule-base simply by adding a new row to the
first structure and then adjusting the weights to appropriate
values.

Now we can explain the function of the first part of the ANN shown
in Fig. \ref{fig3}, which is mathematically described in (\ref{ann1}).
Since we assumed that all inputs and synaptic weights show the confidence degrees and consequently are non-negative
variables, we can conclude that the dot-product of $\mathbf{x}$
and $\mathbf{w}_i^{xv}$ (as well as $\mathbf{y}$ and
$\mathbf{w}_i^{yv}$) indicates the similarity between these two
vectors (or the concepts represented by these membership functions). For example, it is obvious that larger the value of
$\mathbf{w}_i^{xv}\mathbf{x}^T$ larger the similarity between
$\mathbf{w}_i^{xv}$ and $\mathbf{x}$ and therefore larger our confidence degree about the occurrence of the predefined concept $\mathbf{w}_i^{xv}$ at the input variable $x$. This fact provides us with a
new explanation for the task of the thresholding function in ANNs.
To make it clear, consider again the equation relates the output
of neurons of hidden layer to the inputs and synaptic weights:
\begin{equation}\label{ann_thresh}
v_i=f\left(\mathbf{w}_i^{xv}\mathbf{x}^T+\mathbf{w}_i^{yv}\mathbf{y}^T\right),\quad
i=1,2,\ldots,nv.
\end{equation}

In the above equation without using the thresholding function $f$
it is not possible to determine whether the output is caused by
only one of the inputs or both of them. By suitable choice of this
function the output of neuron $i$ is activated only when both
$\mathbf{x}$ and $\mathbf{y}$ are similar to $\mathbf{w}_i^{xv}$
and $\mathbf{w}_i^{yv}$, respectively. Clearly, activation of the
$i$th neuron of hidden layer indicates the simultaneous occurrence
of the concepts defined by $\mathbf{w}_i^{xv}$ and
$\mathbf{w}_i^{yv}$. Hence, the role of
thresholding function $f$ in (\ref{ann_thresh}) is somehow similar
to the role of AND gates in logical circuits. However, since here the probability of having a complete match is very low, hard-thresholding activation function cannot be used. It concludes that in
Fig. \ref{fig3} each row of the crossbar located between input and hidden
layers implements the antecedent part of a fuzzy rule, and the
output of corresponding neuron in hidden layer shows the amount of
activation of that rule (fuzzy min-term) for the given input.

The task of the soft thresholding function, $f$, used in the neurons
of hidden layer can also be interpreted in another way. In Section
\ref{sec_fuzzy} we mentioned that the amount of activation of the
$k$th fuzzy rule is proportional to the amount of overlap between subspaces
$\mathbf{\Sigma}_n^k$ and $\mathbf{\Sigma}_n'$. So, the question
here is: How can we measure the amount of overlap between two
$n$-dimensional subspaces by using the typical integrated circuits
(which are actually two-dimensional devices)? A simple and
efficient answer is that we can measure the similarity at each
dimension separately and then combine the results (by using a
$t$-norm type operator) to obtain the total similarity between
$\mathbf{\Sigma}_n^k$ and $\mathbf{\Sigma}_n'$. It concludes that
the thresholding function, $f$, in ANNs plays the role of $t$-norm
operator in fuzzy inference systems. For example, it can be observed that the ANN shown in Fig. \ref{fig3}
first measures the similarity at each dimension separately (by
calculating $\mathbf{w}_i^{xv}\mathbf{x}^T$ and
$\mathbf{w}_i^{yv}\mathbf{y}^T$ in $x$ and $y$ dimensions,
respectively) and then applies the thresholding function to the
sum of these two values to detect the simultaneous occurrence of
predefined concepts (\textit{i.e.} $\mathbf{w}_i^{xv}$
and $\mathbf{w}_i^{yv}$). As mentioned before, a good method to combine the
similarities obtained at each dimension separately is to use an
operator that  acts somehow as the fuzzy $t$-norm operator. More
precisely, if $a$ and $b$ are two variables of confidence-degree
type (which indicate the similarity between input variables and
the antecedent part of fuzzy rules stored at the rows of crossbar)
the operator $T$ that combines them should have the property:
\begin{equation}\label{t_norm}
\mathrm{if}~ a\le b~\mathrm{then}~T(a,b)\le a.
\end{equation}

\begin{figure*}[!t]
\centering \subfigure[$T_1(a,b)=min(a,b)$]{
\label{fig_op:1} 
\includegraphics[width=3.7cm,height=2.9cm]{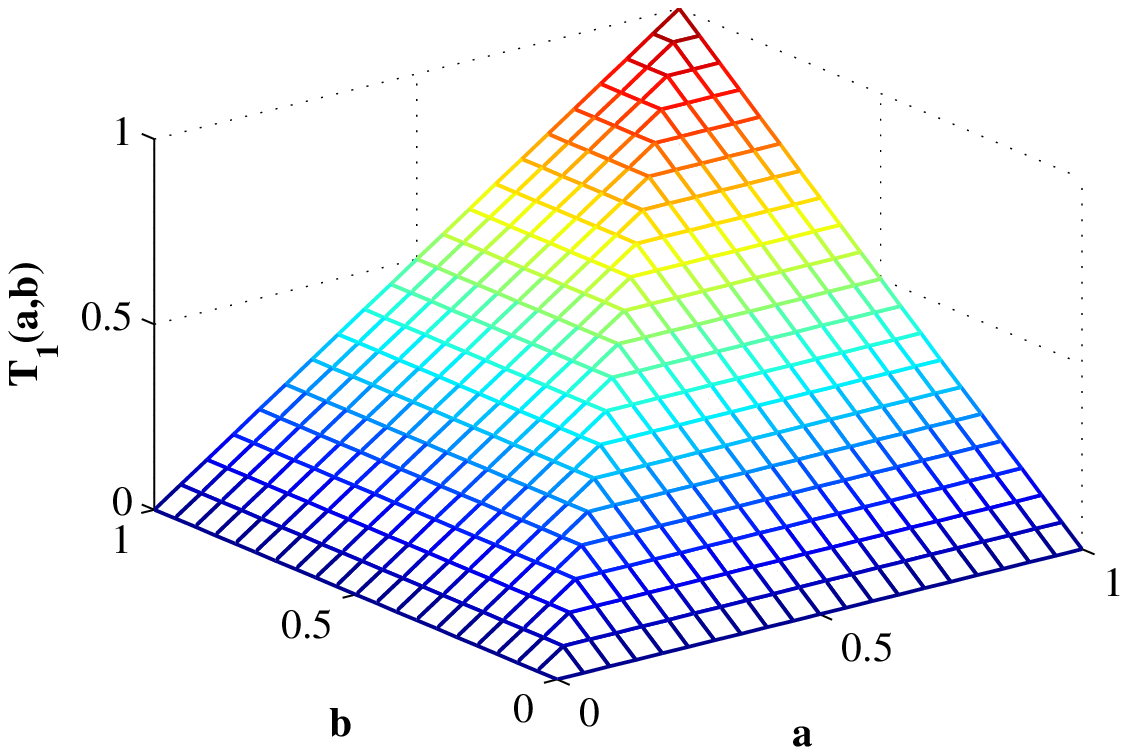}}
\subfigure[$T_2(a,b)=(a+b)^3$]{
\label{fig_op:2} 
\includegraphics[width=3.7cm,height=2.9cm]{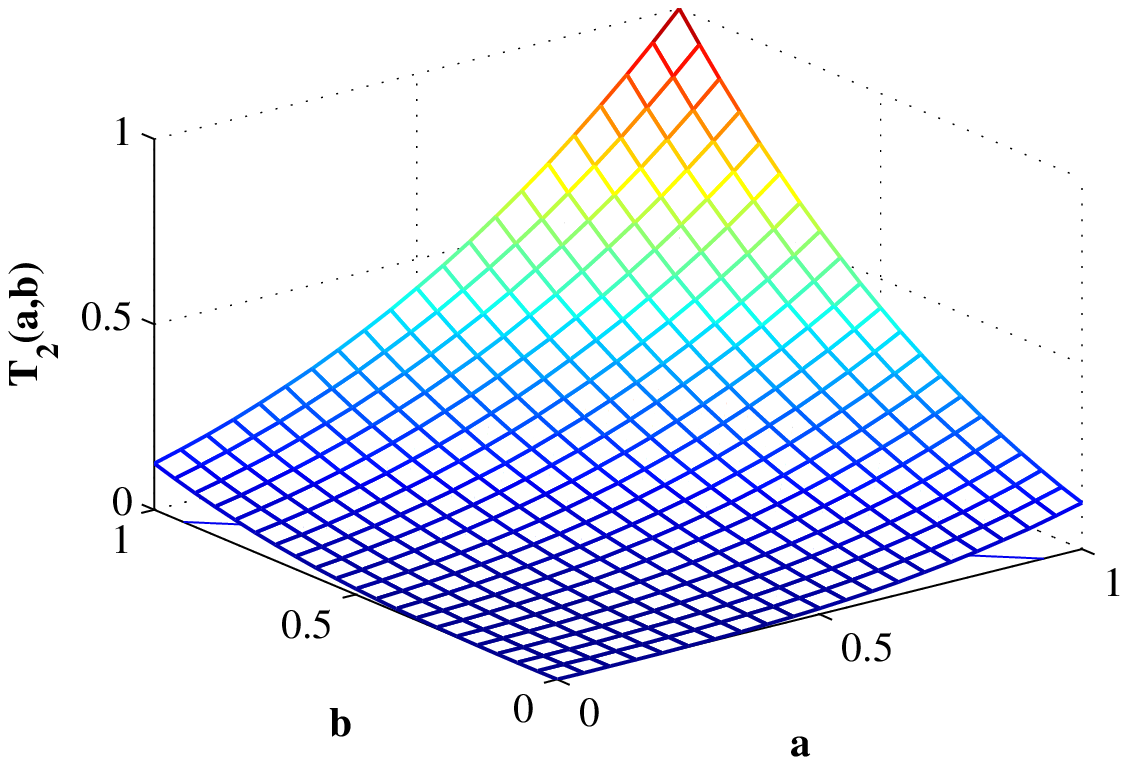}}
 \subfigure[$T_3(a,b)=(a+b)^9$]{
\label{fig_op:3} 
\includegraphics[width=3.7cm,height=2.9cm]{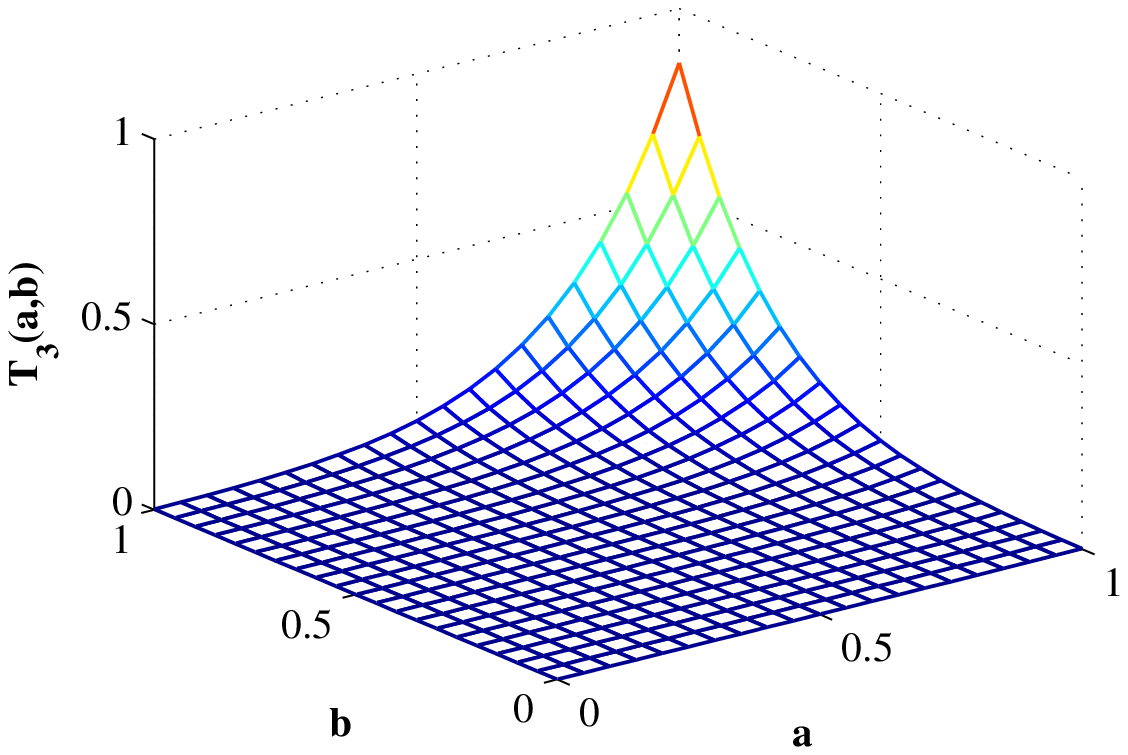}}
 \subfigure[$T_4(a,b)=tansig((a+b)-3)$]{
\label{fig_op:4} 
\includegraphics[width=3.7cm,height=2.9cm]{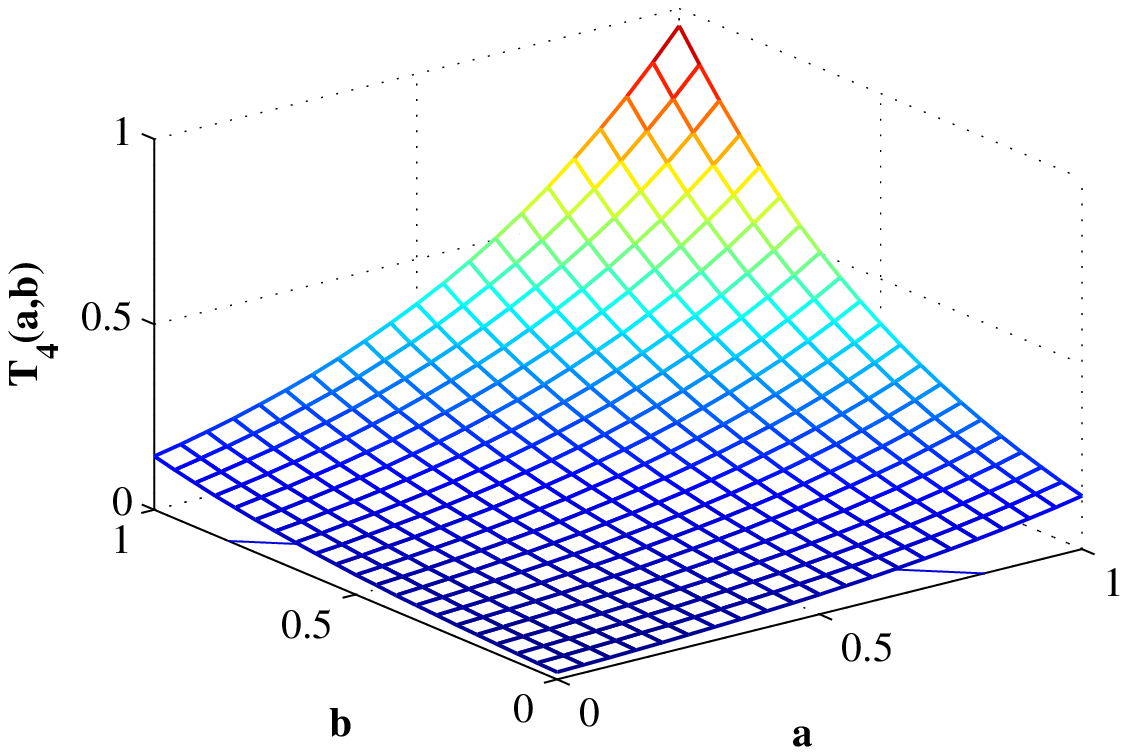}}
 \caption{Comparison between different operators. This figure verifies that the proposed activation function, \textit{i.e.} $(a+b)^p$, acts similar to other $t$-norm operators. In addition, by comparing the shape of operators $T_2$ and $T_4$ it can be said that the activation function of neurons in ANNs to some extend is a $t$-norm operator.}
\label{fig_op} 
\end{figure*}

But, assuming $a=\mathbf{w}_i^{xv}\mathbf{x}^T$ and
$b=\mathbf{w}_i^{yv}\mathbf{y}^T$ the thresholding function given
in (\ref{ann_thresh}) leads to $T(a,b)=f(a+b)$ which, in general,
does not satisfy (\ref{t_norm}). However, by suitable choice of
$f$ it can be observed that this function acts very similar to the
fuzzy $t$-norm operator or the p-input logical AND gate. For example in Fig. \ref{fig_op} we have compared the results
obtained by applying four different operators defined as:
$T_1(a,b)=\min(a,b)$, $T_2(a,b)=(a+b)^3$, $T_3(a,b)=(a+b)^9$, and
$T_4(a,b)=\mathrm{tansig}((a+b)-3)= 2/\left (1+\exp\left
(-2*(a+b-3)\right)\right)-1$ to the operands $a$ and $b$ such that
$0\leq a, b \leq 1$. Note that in this figure output of each
operator is normalized such that it lies between 0 and 1. As it can be seen in this figure, operators $T_2$ and $T_4$ (which is a very common activation function in ANNs) acts similar to fuzzy $t$-norm operator $T_1$. In addition, Fig. \ref{fig_op:3} shows that by increasing the power in the definition of operator $T_2$ this operator can be reduced to the 2-input logical AND gate.

Now, we can study that part of circuit located between hidden and
output layers. In this part of circuit, the weight matrix
$\mathbf{W}^{vz}$ is implemented by using the crossbar structure
similar to previous discussions. It is also assumed that the value
of each neuron at output layer indicates the confidence degree to
a certain point in the universal set of output variable (or the concept represented by that output neuron). Hence,
the values generated by neurons of output layer constitute the
membership function of output fuzzy number. Note that the output of the
ANN shown in Fig. \ref{fig3} can still be connected to the input of
another ANN of this type. The weight matrix $\mathbf{W}^{vz}$
determines which output and to what extent should be activated
when a fuzzy min-term is activated (or equivalently, when certain
concepts  occurred simultaneously at the input). Note that in this
structure we have actually  assumed that the \emph{learning} is the
procedure of making appropriate connections between fuzzy
min-terms and output concepts (but not between input and output
fuzzy concepts). It concludes that the elements of the weight
matrix $\mathbf{W}^{vz}$ can easily be adjusted by applying the
Hebbian learning method (as described in Section \ref{logics}) to the neurons of output and hidden layer.
It means that when for a certain input-output data a neuron in
output layer is activated simultaneous to a neuron in hidden
layer, the synaptic weight connecting these two neurons should be
amplified and the amount of this amplification should be
proportional to the strength of activation of the corresponding
two neurons. Interestingly, in \cite{Farnood2} we showed that this process is also equivalent with the creation of fuzzy relation \cite{leszek} between min-terms and output concepts.  Note that  the membership function of the consequent
part of each fuzzy rule is constructed on a row of the second
crossbar and the final fuzzy output is equal to the weighted sum (aggregation)
of these membership functions, where the weight assigned to each
membership function is proportional to the strength of activation
of the antecedent part of that fuzzy rule.

The interpretation proposed in this section for the function of
ANNs has many advantages to the classical descriptions. First, in
this method all input data, output data and synaptic weights are
non-negative. Second, unlike other training methods (such as
back-propagation) in the proposed structure the appropriate value
of synaptic weights can be obtained without the need to
optimization methods. The main reason for this statement is that
the first crossbar of the ANN shown in Fig. \ref{fig3} implements the
fuzzy min-terms which can be constructed independent of the
definition of final input-output relation.

To sum up, in this section we showed that the operation of ANNs
can be interpreted in a quite different way without applying any
changes to the classical structure. Moreover, we showed that the
two-layer ANN can be considered as a fuzzy inference system with
fuzzy input and fuzzy output. In the next section we present a
method of hardware implementation and training the proposed
neuro-fuzzy computing system, which is obtained by inspiration
from logical circuits, fuzzy logic and ANNs. We will also show
that the proposed neuro-fuzzy system is really effective and can
be used to solve some of the complicated engineering problems.

\section{Realization, training and application of the proposed neuro-fuzzy system}\label{ourmethod}
In this section we explain the hardware implementation, training
and some applications of our proposed nuero-fuzzy system, which
has considerable differences with existing methods. It will also
be shown that the proposed structure has the advantage that can
effectively be designed to deal with massive input-output data.

The overall structure of the proposed neuro-fuzzy computing system is the same as the one depicted in Fig. \ref{fig3}, which
can also be considered as a two-layer network. In this figure,
without any loss of generality, it is assumed that $x$ and $y$ are
scalar inputs and $z$ is the scalar output. Note that this system
is actually a dynamical structure which is incomplete at the
beginning and being more and more completed by training it with the new data. It means
that the hidden layer does not have any neurons before training
and the neurons are generated right after subjecting the system to
input-output training data. For this purpose, first the neurons of
input layer should be divided to few groups such that the number
of groups be equal to the number of input variables. The number of
neurons at each group is, in general, different with others and
depends on the accuracy required to model the variable corresponds
to that group. In fact, at each group one neuron is required  for
any new concept or value that input variable can take. For
example, 101 neurons are required to deal with a variable that
varies between 0 and 10 with the resolution of 0.1. In this case, the first neuron will represent concept ``$x=0$'', the second neuron will represent concept ``$x=0.1$'' and so on. Clearly, by increasing the
number of neurons at input layer any input variable can be
constructed with a desired accuracy. Similarly, one neuron should
be considered at output layer for any distinct value of the output
signal. For the given input, the signal generated at each
neuron of output layer shows the confidence degree of system to
the special value or concept assigned to that neuron. Hence, the output of this
system is actually a fuzzy data which can be converted to a crisp
number by applying any defuzzification method.

It is assumed that the output of each neuron at input layer is
exactly equal to its input (\textit{i.e.}, it has identity activation function), and the output of each neuron at
output layer is equal to the weighted sum of the signals entered to
it (again having identity activation function). As it can be observed in Fig. \ref{fig3} for each of the input
variables a fuzzy set with any desired membership function can be
constructed on each row of the crossbar located between input and
hidden layers. For example, in this figure the fuzzy sets $A_1$
and $A_2$ are constructed on the universal set of $x$ and the
fuzzy sets $B_1$ and $B_2$ are constructed on the universal set of
$y$. It will be shown later that the fuzzy sets constructed on the
rows of this crossbar constitute the antecedent part of if-then
type fuzzy training data.

\subsection{Response of the proposed neuro-fuzzy computing system to the given input}\label{respon}
In this section we assume that the proposed neuro-fuzzy computing
system is designed and fully trained, and we discuss only on
different aspects of computing its output for the given input. The
training procedure will then be discussed in Section \ref{trains}. Note that,
as it will be shown later, the proposed system can be trained
during its ordinary work and the only reason for presenting the
training procedure in another section is for the sake of clarity.

In Fig. \ref{fig3} assume that $x^\ast_i$ is the value of the signal
entered to the $i$th neuron of the group corresponding to variable
$x$, $x_i$ is the concept assigned to the $i$th neuron of this
group, $y^\ast_i$ is the value of the signal entered to the $i$th
neuron of the group corresponding to variable $y$, $y_i$ is the
concept assigned to the $i$th neuron of this group, $nx$ is the
number of neurons used to cover the universal set of the fuzzy
variable $x$, $ny$ is the number of neurons used to cover the
universal set of the fuzzy variable $y$, $z^\ast_i$ is the value
observed at the $i$th neuron of output layer, $z_i$ is the concept
assigned to the $i$th neuron of output layer, $nz$ is the number
of neurons used to cover the universal set of the fuzzy variable
$z$, $v_i$ is the output of $i$th neuron of hidden layer, $N_v^k$
is the number of neurons at hidden layer after applying $k$ set of
input-output training data, $\mathbf{W}^{xv}=\left\{w_{ij}^{xv}
\right\}_{N_v^k\times nx}$ is the weight matrix containing the
coefficients connecting the neurons corresponding to variable $x$
in input layer to the neurons of hidden layer,
$\mathbf{W}^{yv}=\left\{w_{ij}^{yv} \right\}_{N_v^k\times ny}$ is
the weight matrix containing the coefficients connecting the
neurons corresponding to variable $y$ in input layer to the
neurons of hidden layer, and $\mathbf{W}^{vz}=\left\{w_{ij}^{vz}
\right\}_{nz\times N_v^k}$ is the weight matrix containing the
coefficients connecting the neurons of hidden layer to the neurons
of output layer. Using these notations if we denote the input
vectors as $\mathbf{x}^\ast=[x^\ast_1,x^\ast_2,\ldots,
x^\ast_{nx}]$ and $\mathbf{y}^\ast= [y^\ast_1,y^\ast_2
,\ldots,y^\ast_{ny}]$, output of the $i$th neuron at hidden layer
is obtained as
\begin{eqnarray}\label{temp2}
v_i=f\left(\mathbf{w}_i^{xv}{\mathbf{x}^\ast}^T+
\mathbf{w}_i^{yv}{\mathbf{y}^\ast}^T\right) &=&
\left(\mathbf{w}_i^{xv}{\mathbf{x}^\ast}^T+
\mathbf{w}_i^{yv}{\mathbf{y}^\ast}^T\right)^p,\nonumber\\ i &=&1,2,\ldots,
N_v^k,~~p>1,
\end{eqnarray}
where $f(.)$ is the activation function of neurons,
$\mathbf{w}_i^{xv}$ is the $i$th row of $\mathbf{W}^{xv}$ and
$\mathbf{w}_i^{yv}$ is the $i$th row of $\mathbf{W}^{yv}$. The
main difference between (\ref{temp2}) and similar equations
observed in classical ANNs is in the activation function, which is
considered as $f(s)=s^p$ here. In the following we will discuss on
the reason of using this type of activation function in more details.

Consider again the neuro-fuzzy system shown in Fig. \ref{fig3}. As
mentioned before, each row of the first crossbar contains the
membership function of two fuzzy sets, and each fuzzy set is
constructed by equating the synaptic weight stored at each
cross-point to the value of the corresponding membership function
at that point. According to the previous discussions, the data
stored at each row of the first crossbar constitutes a fuzzy
min-term, which approximately refers to a unique combination of
two fuzzy input variables.  It concludes that in practice each distinct
input-output training data can form the antecedent part of an AND-type
fuzzy rule, which is stored on one row of the first crossbar. For
example, the second row of the first crossbar in Fig. \ref{fig3}
implements the antecedent part of the following fuzzy rule:
\begin{equation}
\mathrm{IF} ~x~\mathrm{is}~ A_2~ \mathrm{AND}~y~\mathrm{is}~B_2
~\mathrm{THEN}~z~\mathrm{is}~C_2.
\end{equation}

\begin{figure}[!t]
\centering {
\includegraphics[width=3in,height=3.2in]{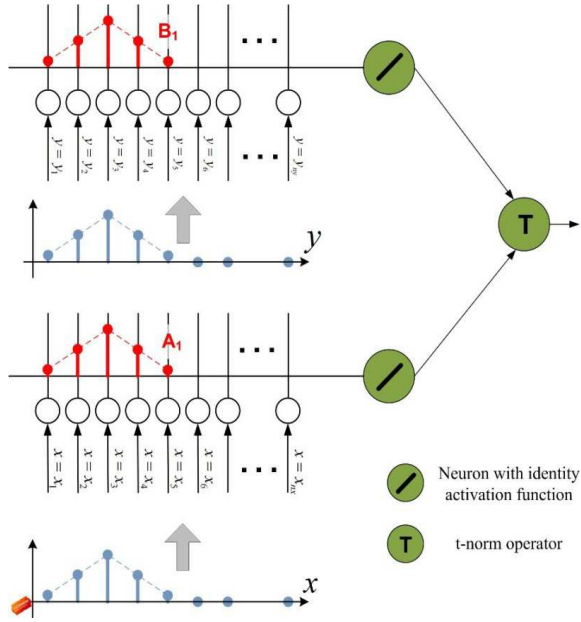}}
\caption{This figure shows how the activation function of neurons can be implemented when the activation function is modeled by a $t$-norm operator. This implementation has this drawback that it depends completely on the number of input variables.}
\label{figtnorm} 
\end{figure}

In order to evaluate the output of each neuron at hidden layer for
the given fuzzy inputs first the similarity between the membership
function of each fuzzy input and the corresponding pattern
(membership function) stored at each row of the crossbar is
determined by calculating the internal product of these two
membership functions. For example, in the second row of crossbar
this procedure is equivalent to the calculation of
$\mathbf{w}_2^{xv} {\mathbf{x}^\ast}^T$ for variable $x$ and
$\mathbf{w}_2^{yv}{\mathbf{y}^\ast}^T$ for variable $y$. Now, in
order to determine the output of the $i$th neuron of hidden layer,
which shows the amount of activation of the corresponding
fuzzy rule, the values obtained for $\mathbf{w}_2^{xv}
{\mathbf{x}^\ast}^T$ and $\mathbf{w}_2^{xv} {\mathbf{y}^\ast}^T$
should be combined using a $t$-norm type operator. In fact, the
operator that combines $\mathbf{w}_2^{xv} {\mathbf{x}^\ast}^T$ and
$\mathbf{w}_2^{yv} {\mathbf{y}^\ast}^T$ should have the property
that generates considerably big outputs only when both of these
two numbers are considerably large. One possible approach for
hardware implementation of such a $t$-norm is to use the structure
shown in Fig. \ref{figtnorm}. The main drawback of this structure, as well as
the logical circuit shown in Fig. \ref{fig1}, is that the number of inputs of
the $t$-norm operator depends on the number of fuzzy inputs of
system. Note that in order to simulate the behavior of brain we
need to develop structures with numerous number of inputs and
outputs where each output is the function of only a few but different number of
inputs. But, if in the structure shown in Fig. \ref{fig3} we use different $t$-norm operators with different number of
 inputs, the resulted hardware will be very
complicated and inefficient. Equation (\ref{temp2}) proposes that
the $t$-norm of $\mathbf{w}_i^{xv} {\mathbf{x}^\ast}^T$ and
$\mathbf{w}_i^{yv} {\mathbf{y}^\ast}^T$ be defined as
$\left(\mathbf{w}_i^{xv} {\mathbf{x}^\ast}^T+ \mathbf{w}_i^{yv}
{\mathbf{y}^\ast}^T\right)^p$, where $p>1$ is an arbitrary
integer. As explained before and showed in Fig. \ref{fig_op}, it can be easily verified that for the values of $p>>1$
the output of each neuron, when both $\mathbf{w}_i^{xv}
{\mathbf{x}^\ast}^T$ and $\mathbf{w}_i^{yv} {\mathbf{y}^\ast}^T$
are large, is much bigger than the case when either
$\mathbf{w}_i^{xv} {\mathbf{x}^\ast}^T$ or $\mathbf{w}_i^{yv}
{\mathbf{y}^\ast}^T$ is large. Although the activation function
defined in (\ref{temp2}) does not satisfy all requirements of a
$t$-norm operator, we will show in Section \ref{simres} that it works very
well in practice. Moreover, it has the advantage that unlike other
activation functions used in classical ANNs does not have any
thresholding or other parameters to tune. Another advantage of
this activation function is that it leads to a structure whose
hidden layer is not changed by increasing the number of the system inputs.

Considering the output of hidden layer as $\mathbf{v}=
[v_1,v_2,\ldots, v_{N_v^k}]$, output of the $i$th neuron at output
layer in Fig. \ref{fig3} is obtained as
\begin{equation}\label{temp3}
z^\ast_i={\mathbf{w}}_i^{vz}\mathbf{v}^T,\quad i=1,2,\ldots,
nz,
\end{equation}
where ${\mathbf{w}}_i^{vz}=[w_{i1}^{vz}, w_{i2}^{vz},
\ldots, w_{i,N_v^k}^{vz}]$ is the $i$th row of
$\mathbf{W}^{vz}$. Equation (\ref{temp3}) can be expressed in
vector form as
\begin{equation}\label{out_mem}
\mathbf{z}^\ast=[z^\ast_1,z^\ast_2,\ldots,
z^\ast_{nz}]=\mathbf{v}\left(\mathbf{W}^{vz}\right)^T.
\end{equation}
where the dimension of  $\mathbf{W}^{vz}$ is $nz\times N_v^k$.

As it can be observed, this part of system acts very similar to
the second layer of classical ANNs and the only difference is that  the
activation function of neurons at the output layer of Fig. \ref{fig3} is of
identical type. In fact, the output vector $\mathbf{z}^\ast$
determines the membership function of the resulted fuzzy output
for the given fuzzy inputs.

\subsection{Training the proposed neuro-fuzzy computing system}\label{trains}
In the previous section we showed that the working procedure of all parts of the proposed
neuro-fuzzy computing system, except the activation function of
neurons at hidden layer, are similar to a classical ANN. However,
what makes the proposed system considerably different and
efficient is the procedure of its training. More precisely, the
proposed system is trained using supervisory method based on the
given input-output data, and as it will be shown later, training
 the synaptic weights or other parameters of the system can be
performed without the need to any optimization method. Without any
loss of generality, consider again the neuro-fuzzy computing
system shown in Fig. \ref{fig3} and assume that the system already has
been subjected to $k$ input-output training data and currently it
has $N_v^k\leq k$ neurons at hidden layer. In the following we discuss on the
method of training the system when it is subjected to a new
training data.

Assume that we are given the $(k+1)$th input-output training data
and our goal is to train the system such that it \emph{learns}
this new data. Note that
since the system under consideration has two inputs and one
output, the new training data must consist of two fuzzy sets as
input variables and a fuzzy set as output variable. But, in the
following, for the sake of simplicity we will assume that the
input training data consists of two fuzzy sets while the output
training data is a crisp number. Denote the fuzzy training inputs
as $\mathbf{x}^\ast$ and $\mathbf{y}^\ast$, and apply them to the
corresponding neurons of input layer (recall that each set of
neurons at input layer covers the universal set of the
corresponding fuzzy input). According to the previous discussions,
applying these fuzzy inputs to system will cause the activation of
the output of each neuron at hidden layer to a certain degree.
Then the output of neurons of hidden layer are combined together
according to (\ref{out_mem}) to form the membership function of
the output variable. Defuzzifying this fuzzy output leads to a
crisp number as the final output. If the difference between
this number and the output training data be less than a predefined
threshold we can conclude that the system already has been trained
with a very similar training data and consequently we do not need
to train the system with this new data again. Two points should be
noted here. First, since we assumed that the output training data
is a crisp number we have to defuzzify the output of our neuro-fuzzy
computing system in order to be able to compare it with the training data. However, in some
applications the output training data is itself a fuzzy concept (hereafter denoted as  $\mathbf{u}^\ast$)
and consequently no defuzzification is required at the output
layer. In this case the difference between the output training
data and output of the neuro-fuzzy computing system can be
measured by calculating, \textit{e.g.}, the internal product of these two
output vectors (membership functions of the two fuzzy numbers). Second, the value of this threshold can easily be
determined according to the accuracy needed by user and without
the need to any optimization. Obviously, decreasing the value of
this parameter will increase the accuracy of system at the cost of
using more fuzzy min-terms (or equivalently, more neurons at
hidden layer).

When the difference between the output generated by the neuro-fuzzy
computing system and the given training output is larger than the
predefined threshold value the system should be trained with the
new training data. In fact, the main reason causes the system not to be able to
generate an accurate output for the given input data is
that it has not been subjected to a similar data before.
In this case first we create a new fuzzy min-term by adding a new
neuron to hidden layer and then we apply appropriate changes to
the synaptic weights stored in the second crossbar to make the
final output closer to the output training data.

The simplest way to create a new fuzzy min-term (to cover the subspace specified by this new data) is to store the
membership functions of the new input data at the cross-points
connecting the neurons of input layer to the new neuron added to
hidden layer. For this purpose we can simply store
$\mathbf{x}^\ast$ and $\mathbf{y}^\ast$ (which are the membership functions describing input data) at the new row added to
the first crossbar. Hence, after adding the neuron number
$N_v^k+1$ to the hidden layer of system we set
\begin{equation}
\mathbf{w}_{N^k_v+1}^{xv}=\mathbf{x^\ast},
\end{equation}
and
\begin{equation}
\mathbf{w}_{N^k_v+1}^{yv}=\mathbf{y^\ast}.
\end{equation}

In this case, the output of this newly added hidden neuron will become active only when the applied input data is close enough to the trained data characterized by $\mathbf{x}^\ast$ and $\mathbf{y}^\ast$. Now, in order to apply appropriate changes to the synaptic weights
stored in the second crossbar (to form connections between min-terms and output concepts) first we calculate the output of
neurons of hidden layer as
\begin{eqnarray}\label{temp4}
v_i=f\left(\mathbf{w}_i^{xv}{\mathbf{x}^\ast}^T+
\mathbf{w}_i^{yv}{\mathbf{y}^\ast}^T\right)=
\left(\mathbf{w}_i^{xv}{\mathbf{x}^\ast}^T+
\mathbf{w}_i^{yv}{\mathbf{y}^\ast}^T\right)^p, \nonumber \\ \quad \quad\quad i=1,2,\ldots,
N_v^k+1,~~p>1.
\end{eqnarray}

Then we update the synaptic weights stored in the weight matrix
$\mathbf{W}^{vz}$ using the Hebbian learning method. In this
method the weight of the synapse connecting two active neurons (one in hidden and another in output layer)   is
amplified proportional to the amount of activation of these two
neurons. More precisely, after calculating the output of neurons at hidden
layer, the synaptic weight connecting the $i$th neuron of hidden
layer to the $j$th neuron of output layer is updated as follows:
\begin{equation}\label{heb1}
w_{ij}^{vz}\leftarrow w_{ij}^{vz}+\alpha t(v_j,u_i),\quad
i=1,\ldots,nz,~j=1,\ldots,N_v^k+1,
\end{equation}
where $u_i$ is the value of the output training data, \textit{i.e.} $\mathbf{u}^\ast$  at the $i$th output neuron, $\alpha$ is the learning
coefficient, and $t$ is the t-norm operator used to implement the
Hebbian learning method. Note that since the latest fuzzy min-term
added to the neuro-fuzzy computing system is exactly the same as
the membership function of fuzzy inputs, output of the $(k+1)$th
neuron of hidden layer would be much bigger than the output of
other neurons at this layer when the system is subjected to the
latest inputs, $\mathbf{x^\ast}$ and $\mathbf{y^\ast}$. Hence, according to (\ref{heb1})
the synaptic weights connecting the $(N_v^k+1)$th neuron of hidden
layer to the neurons of output layer (i.e., the coefficients on
the $(N_v^k+1)$th row of $\mathbf{W}^{vz}$) are mainly affected by
the Hebbian learning method. By repeating the above procedure for
any new input-output training data, the structure becomes more and
more completed. Interesting point in relation to the training of
this system is that, unlike many other training methods, each
input-output data is applied only once and consequently problems
like over-training never occur. Moreover, according to
(\ref{heb1}) this system can be trained by using both fuzzy and
crisp data without the need to any optimization algorithm. 

It may seem that the proposed neuro-fuzzy computing system just
stores the input-output training data and does not perform any
computation. But, by taking into account the possible overlaps
between fuzzy min-terms, it is observed that the size of the data
stored in system can be much smaller than the size of the
input-output training data, specially in dealing with long-term
records. In most of the practical cases the rate of adding new min-terms
to the system is (almost monotonically) decreased by carrying on the
training procedure. Finally, note that another
advantage of the proposed neuro-fuzzy computing system is that
memory and computing units are assimilated together, which is similar
to what has happened in all of the living things.

\section{Hardware implementation of the proposed neuro-fuzzy computing system}\label{hardwa}

In this section, we will discuss on advantages of the proposed
computing system from the hardware implementation point of view.
During the explanation of the proposed method in previous sections
the reader may have recognized that the suggested learning
algorithm is not perfect and can be easily improved in different
ways (e.g., by modifying membership functions of the programmed
fuzzy sets). However, in the following we will show that the main
reason for suggesting this special structure is its simple
hardware implementation. For this purpose first remember that our
main goal is to design a computing system with the ability of
emulating the computing power of human brain. Therefore, by
considering the structural complexity and size of the real neuron,
simplicity of the hardware is of critical importance for its
success. This means that the final system should be easily
expandable by merging basic computing units and it should have a
simple content- or concept-based learning method (similar to
content-based addressable memories with computing ability) which
can be easily mapped into hardware. Hence, using any kind of
optimization method is not allowed evidently since their hardware
implementation is very costly and inefficient even for a small
network.

\begin{figure}[!t]
\centering {
\includegraphics[width=3in,height=2.2in]{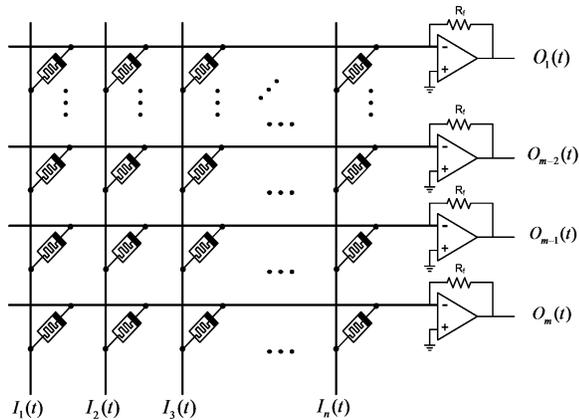}}
\caption{Memristor crossbar-based circuit proposed to do vector to matrix multiplication which can be used for the hardware implementation of the proposed neuro-fuzzy system.}
\label{memcross} 
\end{figure}

Now let us see how the proposed system can be implemented in practice. As stated in Section \ref{respon}, during the ordinary work of system, both parts of the structure shown in Fig. \ref{fig3} perform simple vector to matrix multiplication (see Eqs. \eqref{temp2} and \ref{out_mem}). Note that in this case, the $t$-norm operator or the activation function of neurons is applied to the results obtained from these vector to matrix multiplications. Figure \ref{memcross} shows the proposed circuit to perform vector to matrix multiplication. This circuit consists of a simple memristor crossbar \cite{Strukov2} where each of its rows is connected to the virtually grounded terminal of an operational amplifier that plays the role of a neuron with identity activation function. The memristor crossbar structure has two sets of conductive parallel wires crossing each other perpendicularly such that at each crosspoint a semiconductor device is fabricated between two crossing wires.  In the structure shown in Fig. \ref{memcross} this semiconductor element is a memristive device \cite{williams,Chua2}. Memristive device is a nonlinear element whose resistance (known as memristance) can be tuned by applying a suitable voltage to the device \cite{Alibart2}. Since there is a threshold in the physical model of the device, amplitude of the applied voltage should be larger than this threshold to be able to change the state (and consequently, the memristance) of the device \cite{Lehtonen2}. Hence,  assuming that the amplitude of the  voltages applied to the circuit of Fig. \ref{memcross} is below the threshold of memristive devices,  output of the $i$th neuron (operational amplifier), $O_i$, can be written as:
\begin{equation}\label{crosseq}
O_i=-\sum_{j=1}^n\left(\frac{R_{f}}{M_{ij}}\right)I_j=-\sum_{j=1}^nR_fG_{ij}I_j
\end{equation}
where $M_{ij}$ and $G_{ij}$ are the memristance and memductance (inverse of memristance) of the memristive device located in the crossing point of the $i$th row and the $j$th column of the crossbar, $R_f$ is the feedback resistor of operational amplifiers and $I_j$ is the input voltage applied to the $j$th column of the crossbar. Note that since the memristance of memristive devices is not changing during this computation, in Eq. \eqref{crosseq} they have been treated as ordinary resistors. By comparing Eqs. \eqref{crosseq} and \eqref{temp2} (or Eqs. \eqref{crosseq} and \eqref{out_mem}) it becomes clear that this structure is a perfect circuit to implement the proposed method. For this purpose, connection weight matrices (which contain the membership function of corresponding fuzzy sets) should be stored as a memductance of memristive devices in crosspoints of the crossbar. On the other hand, each neuron should compute the weighted sum of its inputs and then apply the activation function to the result. This can be done by applying a suitable non-linear function, \textit{e.g.} $f(\cdot)=(\cdot)^p$, to the output of operational amplifiers in Fig. \ref{memcross}. Finally, it is evident that the computing structure shown
in Fig. \ref{fig3} can be implemented by series connection of two of
these memristor crossbar structures (\textit{i.e.}, connecting the
outputs of one of these circuits directly to the inputs of another
one).

Now, consider the learning process described in Section \ref{trains}. As explained in that section, the learning process of the proposed method consists of two different phases: (i) creation of min-terms (storing fuzzy sets) on the first crossbar (which connect the neurons of input layer to the neurons of hidden layer) and (ii) updating the connection weights stored in crosspoints of the second crossbar (which connect the neurons of hidden layer to the neurons of output layer). In order to add a new min-term to the sample of the structure shown in Fig. \ref{memcross}, first a new row should be added to the first crossbar of this structure. For this
purpose from the beginning we can reserve some of the rows of the
first crossbar for storing upcoming data. In this case adding a
new row is equivalent to tuning the memristance of the memristors
located on an unused row to suitable values. In order to store
weights on the newly added row (or equivalently, storing the
membership function of fuzzy input sets) we interpret the value of
the membership function at each point as a voltage signal and then
we apply it to the corresponding column of the crossbar. Now, by
grounding the new row while other rows are connected to a high
impedance, current will pass through the memristors connected to
this row. Assuming that all of these memristors initially have a
same memristance and considering the fact that the amount of
electrical current passing through a memristor is proportional to
the amplitude of the voltage applied to it, it can be concluded
that the conductance of each memristor on the newly added row is
proportional to the amplitude of the voltage applied to the
corresponding column (or equivalently, to the membership function
of fuzzy input at that column). This means that only by applying membership functions to columns
of the first crossbar while newly added row is grounded and then
waiting for specific time, a new min-term corresponding to the
input training data automatically will be added to the crossbar.
Note that several methods have been proposed so far to change the
memristance of specified memristors in a crossbar without altering
the memristance of other semi-selected memristors \cite{Strukov2}.

\begin{figure}[!t]
\centering {
\includegraphics[width=3.2in,height=2.2in]{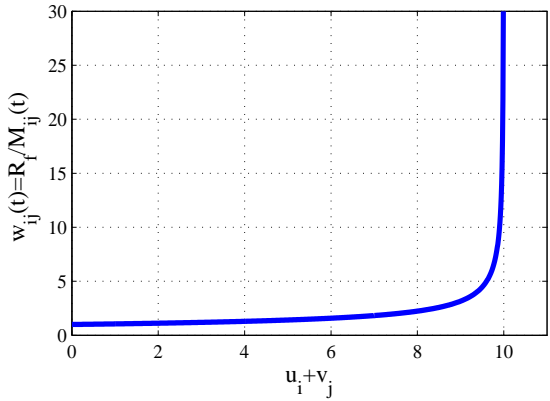}}
\caption{This figure shows how the weight stored in the memristor typically changes versus the applied voltage during the learning process. From this figure it can be figured out that a single memristor acts as a $t$-norm operator. It changes the stored value (memristance of the memristor) significantly when the voltage across the device is big (so the probability that both voltages at terminals of memristor have big values is high as well). In this simulation the HP model \cite{williams} is used to simulate memristor and voltages are applied to memristor for about 0.05 second.}
\label{memrespp} 
\end{figure}

Now, let us consider the problem of hardware implementation of the second phase of the proposed learning method on the structure shown in Fig. \ref{memcross}, which is used to implement the second crossbar in Fig. \ref{fig3} (from hidden layer to output layer). In fact, here we want to update the corresponding weights (\textit{i.e.} memductance of memristors) based on the given training data. It is concluded from Eq. \eqref{heb1} that for any given training data, memductance (connection weight) of the memristor connecting a neuron of hidden layer to a neuron of output layer should be changed such that the
amount of this change be proportional to the sum of the firing
strength of these two neurons. In order to implement this updating
method in the memristive structure shown in Fig. \ref{memcross}, first we
inject the input training data to system and let the neurons of
hidden layer generate their own output signals (see Eq. \eqref{temp2}). Then, we interpret the value of membership function of the output fuzzy training data (or $\mathbf{u^\ast}$) at any point as a voltage signal and apply the negative of this voltage to its corresponding column of the crossbar. In this case, the current passing through the memristor connecting a hidden neuron to an output neuron will be proportional to the voltage dropped across this memristor, which is equal to the sum of the absolute value of the voltages applied to row and column of the crossbar that this memristor is located between them. For example, for the memristor located in the crossing point of the $i$th row and the $j$th column of the crossbar this voltage will be equal to $u_i+v_j$. Now, application of this voltage will cause the memristance(memductance) of the memristor to decrease(increase) which will increase the weight $w_{ij}=\frac{R_f}{M_{ij}(t)}$ stored in this memristor (so the relation expressed in Eq. \eqref{heb1}). Figure \ref{memrespp} shows how the weight stored in a typical memristor, \textit{i.e.} $w_{ij}=\frac{R_f}{M_{ij}(t)}$, changes versus the amplitude of the voltage $u_i+v_j$ when it is applied to the device for a specific period of time. In this figure, $R_f$ and the initial memristance of the memristor are considered equal to $R_{off}$ (the maximum memristance that the memristor can have). 

By comparing Figs. \ref{memrespp} and \ref{fig_op} it can be inferred that each memristor somehow applies a $t$-norm operator to the two voltages connected to its terminals (so the function $t(u_i,v_j)$ in Eq. \eqref{heb1}): when both voltages have high values (\textit{i.e.} the neurons are fired simultaneously), the weight stored in the memristor is increased much more than the case in which only one of these neurons is fired. This means that at the end of this learning procedure, we will see a strong connection only between those hidden and output neurons that usually fire simultaneously. To summarize, to carry out this learning process we should only apply training data to input and output terminals of the structure and wait some period of time. This will cause the memristance of memristors to change based on the Hebbian learning rule similar to what we had in Eq. \eqref{heb1}. To conclude, by using two of these memristor crossbar structures, connecting them to each other correctly and managing the amplitude and timing of the applied voltages corresponding to training data, the proposed neuro-fuzzy system or any rule-based fuzzy inference method can be simply constructed.

\section{Simulation results}\label{simres}
\begin{figure*}[!t]
\centering \subfigure[$g_1$]{
\label{fig:5-7} 
\includegraphics[width=5cm,height=4cm]{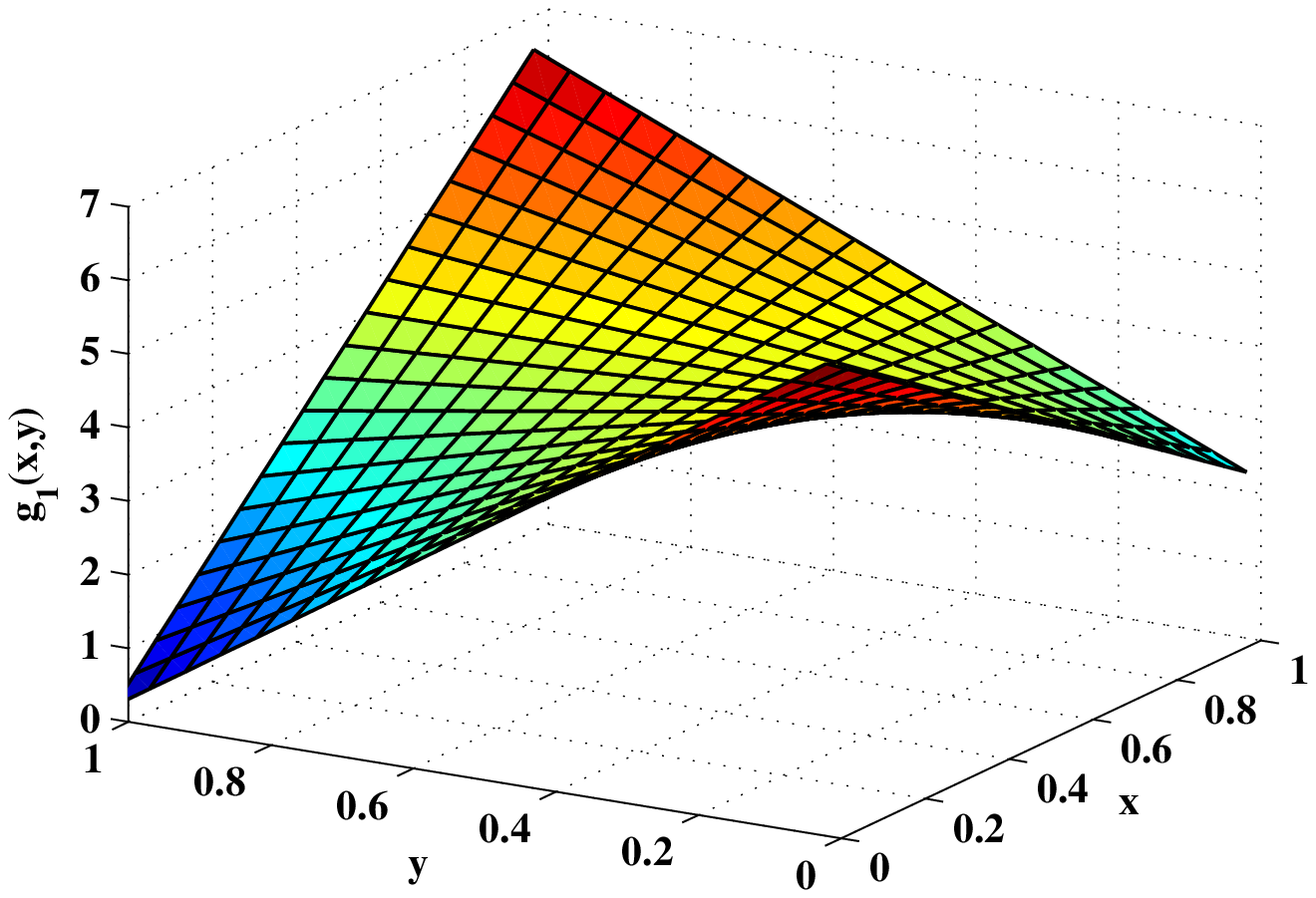}}
\subfigure[$g_2$]{
\label{fig:5-8} 
\includegraphics[width=5cm,height=4cm,origin=c]{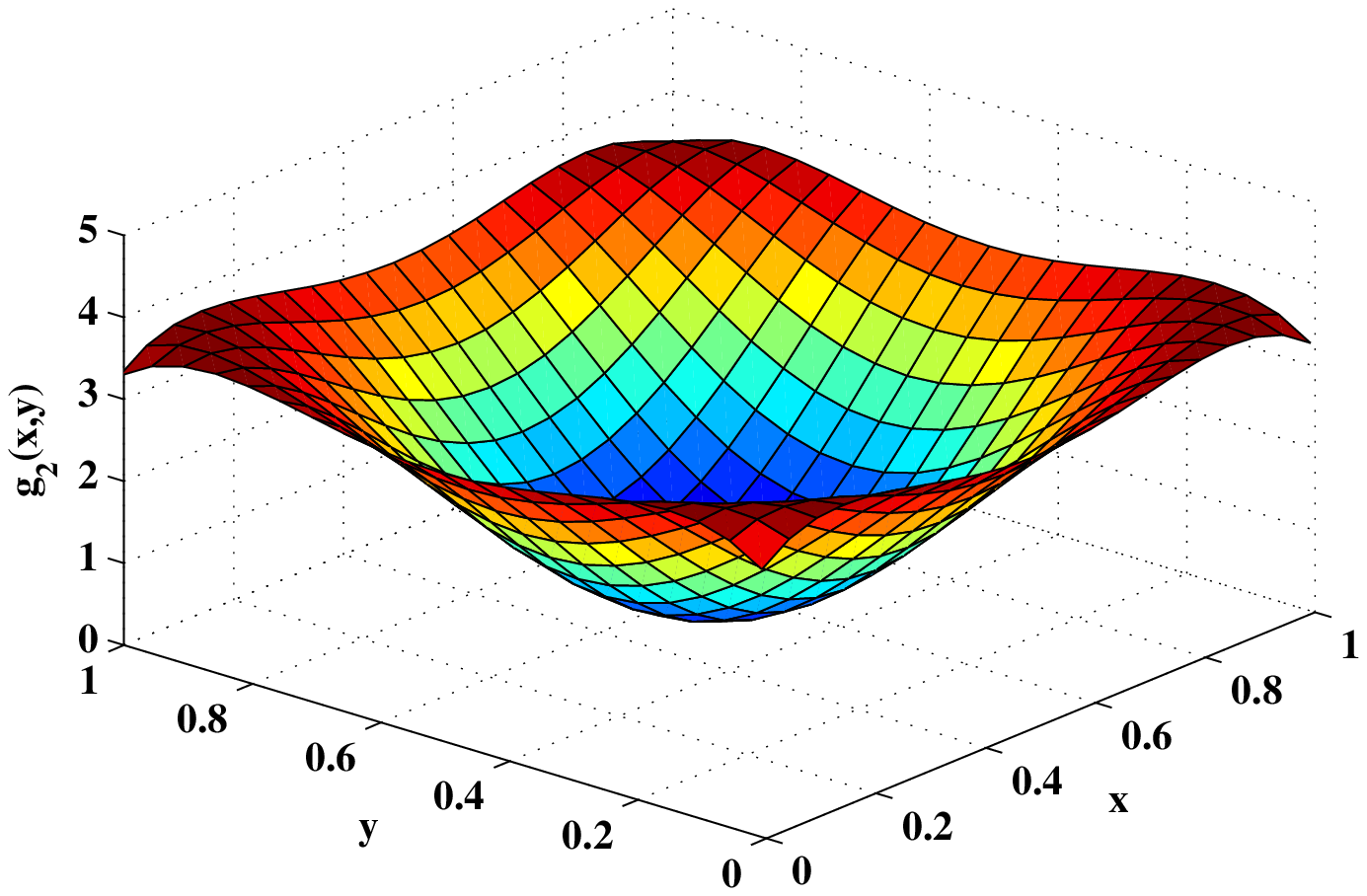}}
\subfigure[$g_3$]{
\label{fig:5-9} 
\includegraphics[width=5cm,height=4cm,origin=c]{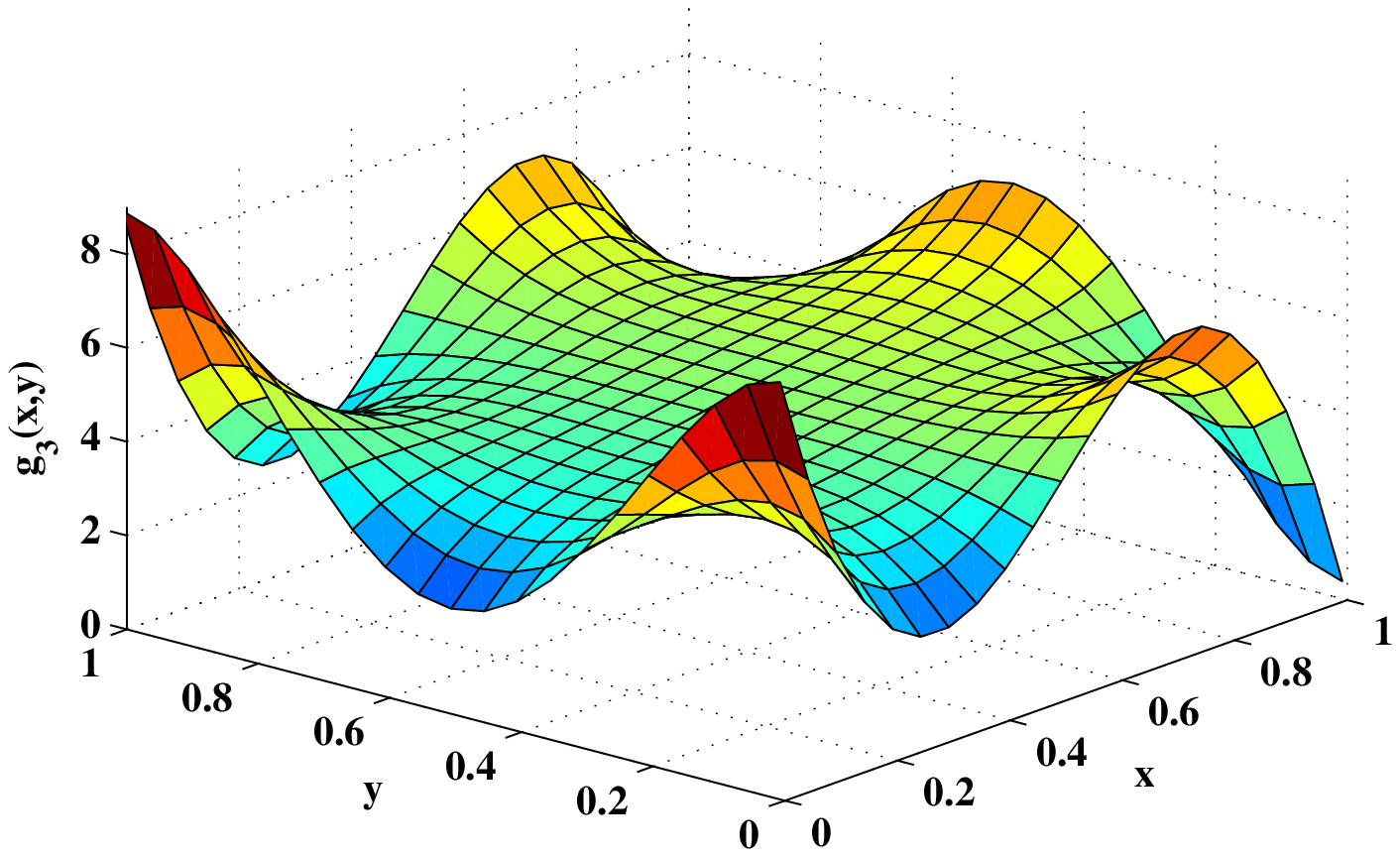}}
\subfigure[$g_4$]{
\label{fig:5-10} 
\includegraphics[width=5cm,height=4cm,origin=c]{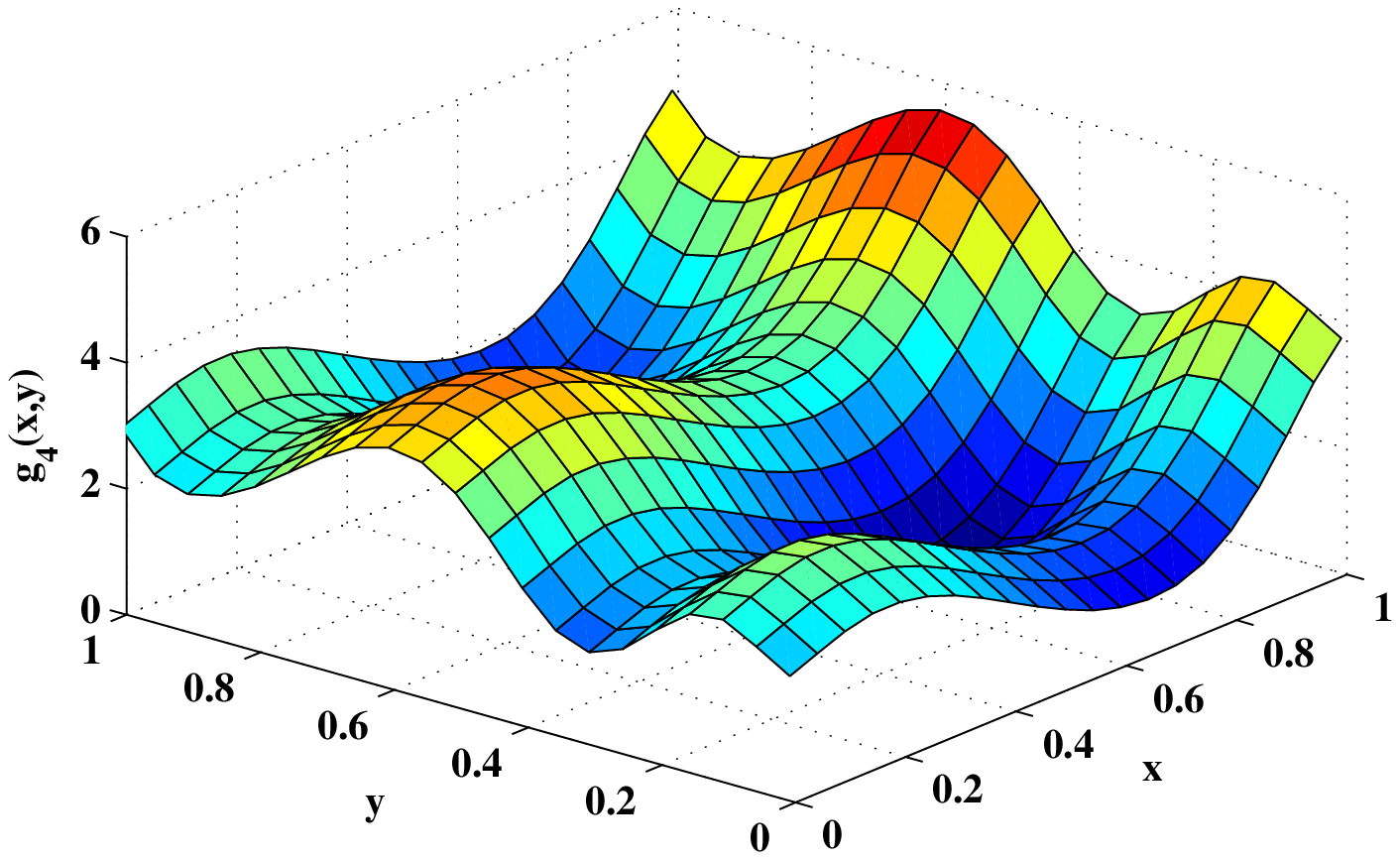}}
\subfigure[$g_5$]{
\label{fig:5-11} 
\includegraphics[width=5cm,height=4cm,origin=c]{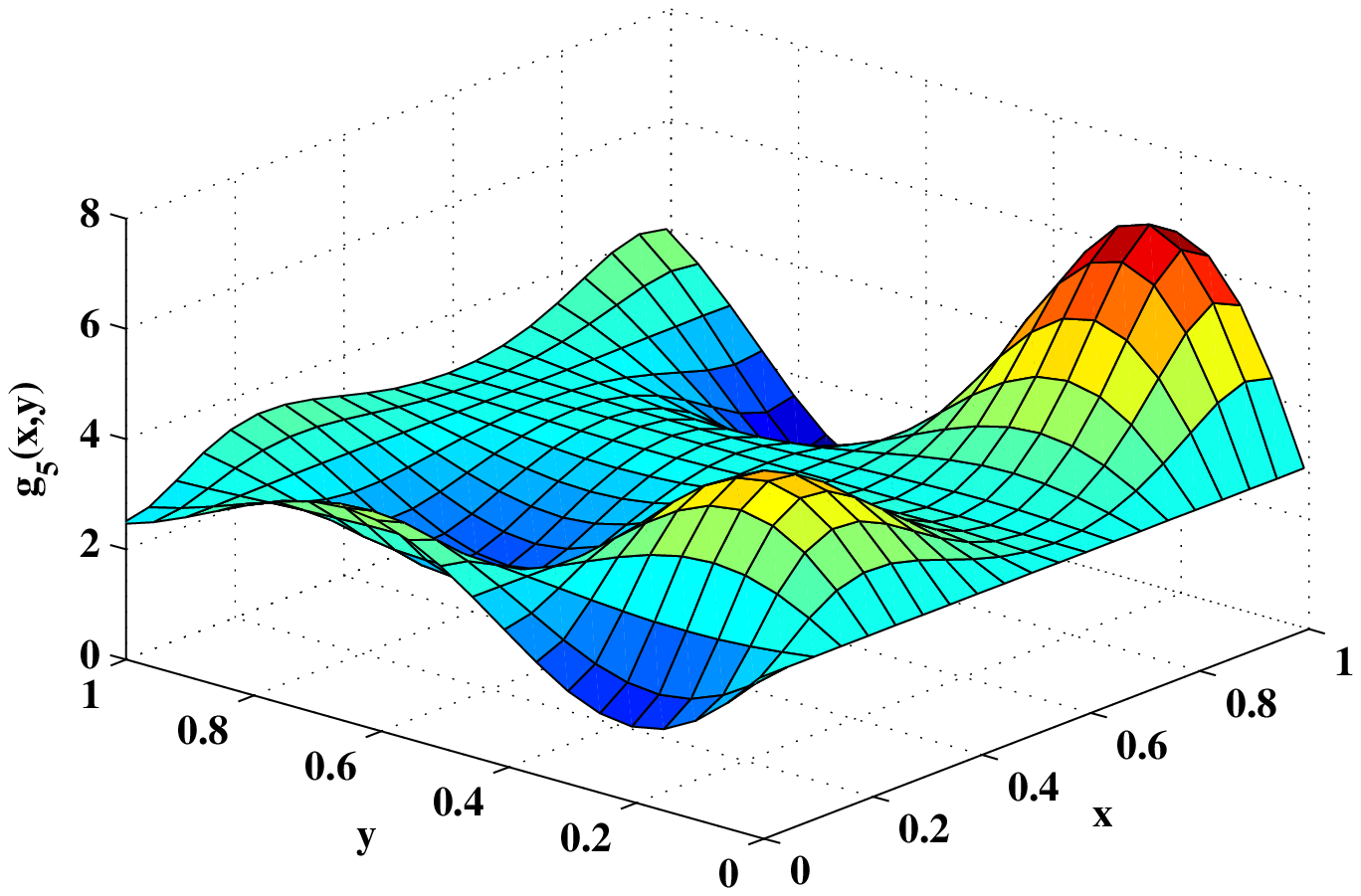}}
\caption{Graph of functions $g_1$ to $g_5$.}
\label{fig5-7and8and9and10and11} 
\end{figure*}

In this section we show the high potential of the proposed
neuro-fuzzy computing system for solving different engineering
problems in the field of modeling and classification. One
important capability of the proposed system is to model
multi-variable mathematical functions. To show this, consider the
functions:
\begin{equation}\label{g1}
g_1(x,y)=10.391((x-0.4)(y-0.6)+0.36),
\end{equation}
\begin{equation}
g_2(x,y)=24.234\left(r^2(0.75-r^2)\right),\ \  r^2=(x-0.5)^2
+(y-0.5)^2,
\end{equation}
\begin{eqnarray}
g_3(x,y)&=&42.659\left(0.1+\widetilde{x} \left(0.05+ \widetilde{x}^4
-10\widetilde{x}^2\widetilde{y}^2 +5\widetilde{y}^4\right)
\right),\nonumber\\&& \widetilde{x}=x-0.5,\quad \widetilde{y}= y-0.5,
\end{eqnarray}
\begin{eqnarray}
g_4(x,y)&=&1.3356\left(1.5(1-x)+e^{2x-1}\sin(3 \pi(x-0.6)^2)\right)\nonumber\\
&+&1.3356\left(e^{3(y-0.5)} \sin(4\pi(y-0.9)^2)\right),
\end{eqnarray}
\begin{equation}\label{g5}
g_5(x,y)=1.9(1.35+e^x\sin(13 (x-0.6)^2) e^{-y} \sin (7y)),
\end{equation}
which are proposed by Hwang et al. \cite{Hwang1994} to study the learning and
modeling capability of systems (in all of the above
functions it is assumed that $x,y\in [0,1]$). In the following we
use the structure shown in Fig. \ref{fig3} to model each of these
functions, which are also shown in Fig. \ref{fig5-7and8and9and10and11}.

\begin{figure}
\centering{
\includegraphics[width=8cm,height=2.2cm]{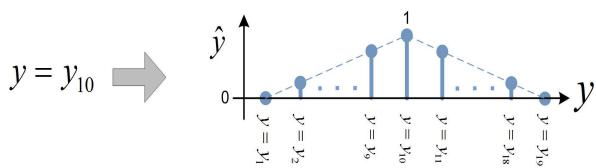}}
\caption{Fuzzification process used to convert crisp training data to their corresponding fuzzy numbers.}
\label{fig5-12}
\end{figure}

In each simulation the neuro-fuzzy computing system is subjected
to 225 training data and 10,000 test data, which are randomly
selected from the space of input variables. Moreover, power of the
activation function of neurons at hidden layer (i.e., the value of
$p$ in (\ref{temp2})) and the value of $\alpha$ in (\ref{heb1})
are considered equal to 7 and 0.0005, respectively (these values
are obtained by a simple trial and error and it is observed during
numerical simulations that the final results are not so sensitive
to the values assigned to these parameters). The Fraction
of Variance Unexplained (FVU) performance index defined as follows
\cite{Hwang1994}
\begin{eqnarray}
FVU=\frac{\sum_{i=1}^{10000}(g(x_i,y_i)-\widehat{g}(x_i,y_i))^2 }
{\sum_{i=1}^{10000}(g(x_i,y_i)-\overline{g})^2 },\nonumber \\ \quad
\overline{g}=\frac{1}{10000} \sum_{i=1}^{10000}g(x_i,y_i),
\end{eqnarray}
is used to evaluate the modeling accuracy of the resulted neuro-fuzzy
computing systems, where $g(x_i,y_i)$ and $\widehat{g}(x_i,y_i)$
are the values generated by function itself and the proposed neuro-fuzzy
computing system, respectively. Table \ref{table1} summarizes the
simulation results. Note that since the original training data
consist of crisp numbers, in all simulations each input data is
subjected to a fuzzification as shown in Fig. \ref{fig5-12} before entering
to the system. In this fuzzification process, the support set of the resulted fuzzy number should be chosen proportional to the number of available training data and the accuracy required for modeling the function. By increasing of the number of training data, smaller support set can be chosen to reach higher modeling accuracy. Table \ref{table1} shows that, for example, 100 neurons are
used to cover the universal set of $x$ and $y$, and 116 neurons
are used to cover the universal set of $z$, when modeling of
$g_1$ is aimed. According to the domain of definition and the
range of $g_1$ it concludes that the variables $x$, $y$, and $z$
are modeled with the maximum accuracy of 0.01, 0.01, and 0.06,
respectively.


\begin{table*}[ht]
\small
\caption{Accuracies of the models optimized using the proposed method.} 
\centering 
\begin{tabular}{l p{2.2cm} p{2.2cm} p{2.2cm} l l p{2.1cm}} 
\hline\hline 
Function & \# of neurons for variable $x$ & \# of neurons for variable $y$ & \# of neurons for variable $z$ & Threshold value & $FVU$ & \# of constructed min-terms \\[0.5ex]
\hline 
$g_1$ &100 & 100 & 116 & 0.2 & 0.067 & 77 \\ 
$g_2$ &100 & 100 & 69 & 0.1 & 0.044 & 161 \\
$g_3$ &100 & 100 & 143 & 0.1 & 0.263 & 140 \\ 
$g_4$ &100 & 100 & 105 & 0.2 & 0.087 & 123 \\
$g_5$ &100 & 100 & 126 & 0.15 & 0.09 & 130 \\[1ex] 
\hline 
\end{tabular}
\label{table1} 
\end{table*}

Table \ref{table2} summarizes the results obtained by modeling
$g_1$,...,$g_5$ with other methods. The first group of the results
presented in this table corresponds to the modeling of these
functions by the ANNs trained using back-propagation and Projection Pursuit Learning (PPL)
method (under different conditions and structures). The second
group of the results presented in Table \ref{table2} corresponds to the
method proposed by Kwok and Yeung \cite{Kwok1997} for training the
coefficients of the new hidden units added to a dynamical network based
on using different cost functions (\textit{i.e.}, $S_1$, $\sqrt{S_1}$,
$S_2$, $\sqrt{S_2}$, $S_3$, $\sqrt{S_3}$, $S_{cascor}$,
$S_{fujita}$, and $S_{sqr}$ as defined in \cite{Kwok1997}). The third group
of results in Table \ref{table2} is obtained by using the method proposed by
Ma and Khorasani \cite{Ma2005}, which is similar to the previous method
with the difference that instead of using different cost functions
for training, different Hermite polynomial activation functions are applied to the
neurons of hidden layer. Murakami and Honda \cite{Murakami2007} studied the modeling ability of the Active Learning Method (ALM) and used it for pattern-based information processing. The ALM divides the input space into several partitions and then models the function in each of these partitions through a simple pattern. Finally, the last group of
results in Table \ref{table2} corresponds to the modeling of functions
using the ANFIS method \cite{Jang1993}.

\begin{table*}[t]
\small
\caption{Comparing the accuracies of the models optimized using different algorithms based on FVU criterion \cite{Murakami2007}.}
\centering
\begin{tabular}{c p{10cm} c c c c c} 
\hline\hline
Ref. & Model &\multicolumn{5}{c}{Function}\\
\cline{3-7}
  &   & $g_1$& $g_2$ & $g_3$ & $g_4$ & $g_5$\\
\hline 
\cite{Hwang1994} & BPL based on Gauss-Newton method (5 hidden units) & 0.001 & 0.065 & 0.506 & 0.080 & 0.142\\
 &  {BPL based on Gauss-Newton method (10 hidden units)} & 0.001 & 0.002 & 0.183 & 0.003 & 0.021\\
 &  {PPL supersmoother (3 hidden units)} & 0.000 & 0.010 & 0.355 & 0.021 & 0.135\\ 
 &  {PPL supersmoother (5 hidden units)} & 0.000 & 0.007 & 0.248 & 0.000 & 0.028\\ 
 &  {PPL Hermite (3 hidden units)} & 0.000 & 0.009 & 0.075 & 0.001 & 0.049\\ 
 &  {PPL Hermite (5 hidden units)} & 0.000 & 0.000 & 0.000 & 0.001 & 0.015\\ [1ex]
 \cite{Kwok1997} &  {CFNN with $S_1$} & 0.021 & 0.029 & 0.269 & 0.036 & 0.121\\ 
 &  {CFNN with $\sqrt{S_1}$} & 0.011 & 0.028 & 0.247 & 0.037 & 0.111\\ 
 &  {CFNN with $S_2$} & 0.095 & 0.426 & 0.547 & 0.636 & 0.610\\ 
 &  {CFNN with $\sqrt{S_2}$} & 0.024 & 0.031 & 0.275 & 0.031 & 0.134\\ 
 &  {CFNN with $S_3$} & 0.003 & 0.020 & 0.306 & 0.027 & 0.160\\ 
 &  {CFNN with $\sqrt{S_3}$} & 0.003 & 0.018 & 0.288 & 0.030 & 0.167\\ 
 &  {CFNN with $S_{cascor}$} & 0.025 & 0.027 & 0.265 & 0.031 & 0.121\\ 
 &  {CFNN with $S_{fujita}$} & 0.004 & 0.047 & 0.444 & 0.070 & 0.246\\ 
 &  {CFNN with $S_{sqr}$} & 0.007 & 0.038 & 0.573 & 0.185 & 0.294\\[1ex] 
\cite{Ma2005} &  {Standard CFNN with sigmoidal activation functions (10 hidden units)} & 0.048 & 0.097 & 0.551 & 0.073 & 0.206\\ 
 & {Proposed CFNN with Hermite polynomial activation functions (10 hidden units)} & 0.031 & 0.027 & 0.197 & 0.076 & 0.095\\ 
 &  {Standard CFNN with sigmoidal activation functions (20 hidden units)} & 0.043 & 0.048 & 0.303 & 0.050 & 0.111\\ 
 &  {Proposed CFNN with Hermite polynomial activation functions (20 hidden units)} & 0.026 & 0.019 & 0.082 & 0.027 & 0.039\\ [1ex]
\cite{Murakami2007} &  {ALM with 6 partitions} & 0.014 & 0.031 & 0.153 & 0.057 & 0.076\\ 
 &  {ALM with 7 partitions} & 0.015 & 0.027 & 0.132 & 0.060 & 0.062\\ 
 &  {ALM with 8 partitions} & 0.021 & 0.032 & 0.129 & 0.061 & 0.063\\ 
 &  {ALM with 9 partitions} & 0.027 & 0.035 & 0.122 & 0.067 & 0.064\\[1ex]
 \cite{Jang1993} &  {ANFIS with 9 rules} & 0.000 & 0.002 & 0.033 & 0.008 & 0.089\\ [1ex]
 \hline
\end{tabular}
\label{table2}
\end{table*}


Comparing the modeling errors of $g_1$ in Tables \ref{table1} and \ref{table2}
leads to the fact that the proposed neuro-fuzzy computing system is less
effective than other methods for the modeling of approximately
linear functions such as $g_1$. The main reason for this problem
is that the proposed system divides the space of input variables into several overlapping subspaces (min-terms) and tries to model the given function in each of these subspaces by a single IF-THEN rule. However, it is a well-known fact that by the aggregation of this kind of rules (which has a simple fuzzy set in their consequent part), it is very difficult to model a linear function. It is also concluded from Table \ref{table1} that in all
cases a high percent of training data is stored in the crossbar.
The reason for this problem is that since the number of training data is not
large enough, most of the input-output training pairs contain a
new information and tend to constitute a new fuzzy min-term. It
will be shown later that this problem can be removed simply by
increasing the number of input-output training pairs. 

Here, it is worth to emphasize on three important notes. Firstly, although some other neuro-fuzzy systems such as ANFIS can outperform the proposed system, but they have the disadvantage that after their training the resulted fuzzy sets and rules
conceptually have almost no meaning. Secondly, except the ALM, all other methods are relied on optimization methods, which means that they do not have biological support and they cannot be implemented in large scale. In addition, they suffer from the very high computational cost. Finally, it should be noted that unlike almost all other methods, in the proposed computing system each training data is presented to the system only once. 



Table \ref{table3} shows the effect of increasing the number of training
data on the accuracy and the number of constructed fuzzy min-terms of the
resulted system. Note that in each case the number of fuzzy
min-terms is equal to the  number of dissimilar training data
sufficient to construct the system. It is concluded from Table \ref{table3} that
the proposed neuro-fuzzy system can effectively model the
functions under consideration by considering only some of the most
important fuzzy min-terms (training data). Another observation is that the
accuracy of system can considerably be increased by increasing the
number of input-output training data. Finally, note that since in
all simulations parameters of the proposed neuro-fuzzy computing
system  are obtained by trial and error (and consequently, are not
optimal), it is expected that better results can be achieved by
using optimally selected parameters.

\begin{table}[ht]
\small
\caption{Accuracies of the models optimized using the proposed method with different numbers of training data.} 
\centering 
\begin{tabular}{c| c c| c c| c c} 
\hline\hline 
Function & \multicolumn{2}{c}{$g_1$}&\multicolumn{2}{c}{$g_3$}&\multicolumn{2}{c}{$g_5$}\\
\hline
\# of training data &400& 700&400&700&400&700\\
\hline FVU & 0.026& 0.021&0.153&0.117&0.058&0.036\\
\# of min-terms&194&238&216&245&189&229\\[1ex]
\hline
\end{tabular}
\label{table3} 
\end{table}

\begin{figure*}[!t]
\centering \subfigure[Data set \#1]{
\label{fig:5-13} 
\includegraphics[width=3.6cm,height=3.4cm]{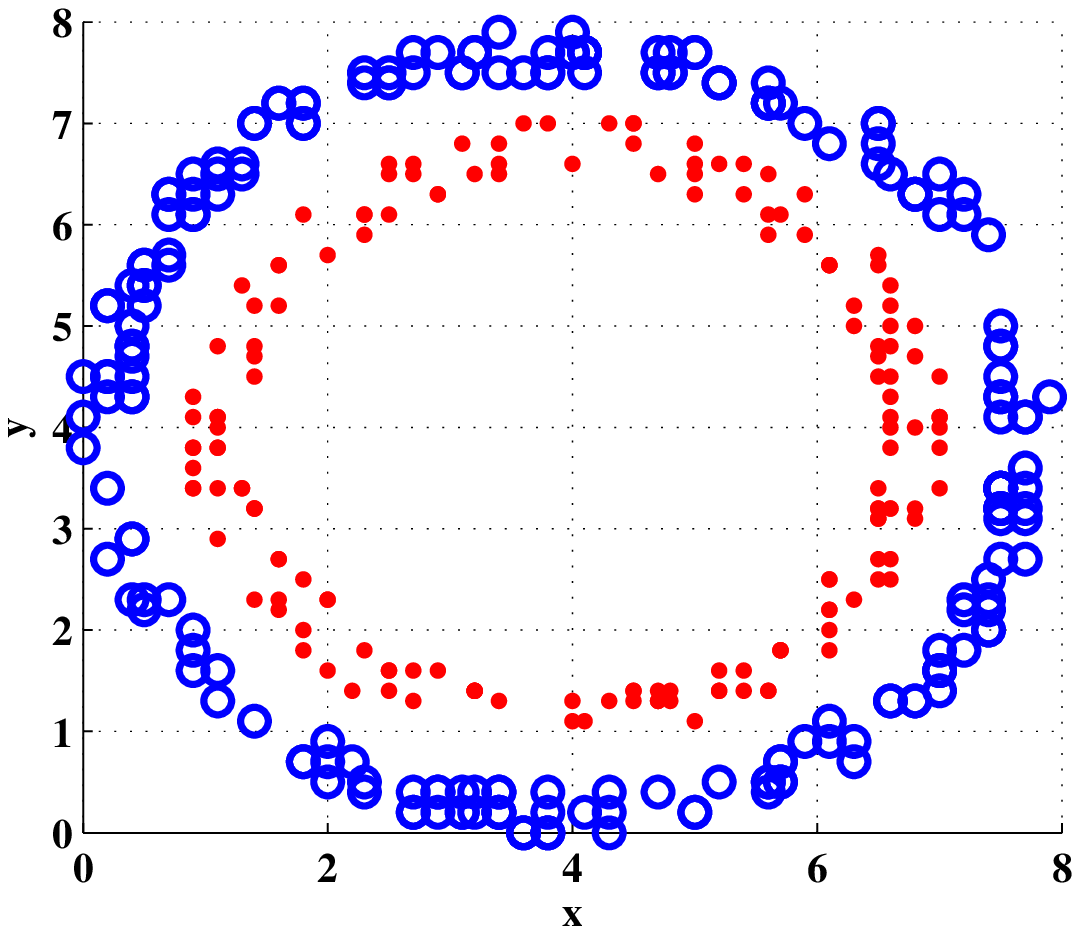}}
\subfigure[Data set \#2]{
\label{fig:5-14} 
\includegraphics[width=3.6cm,height=3.4cm,origin=c]{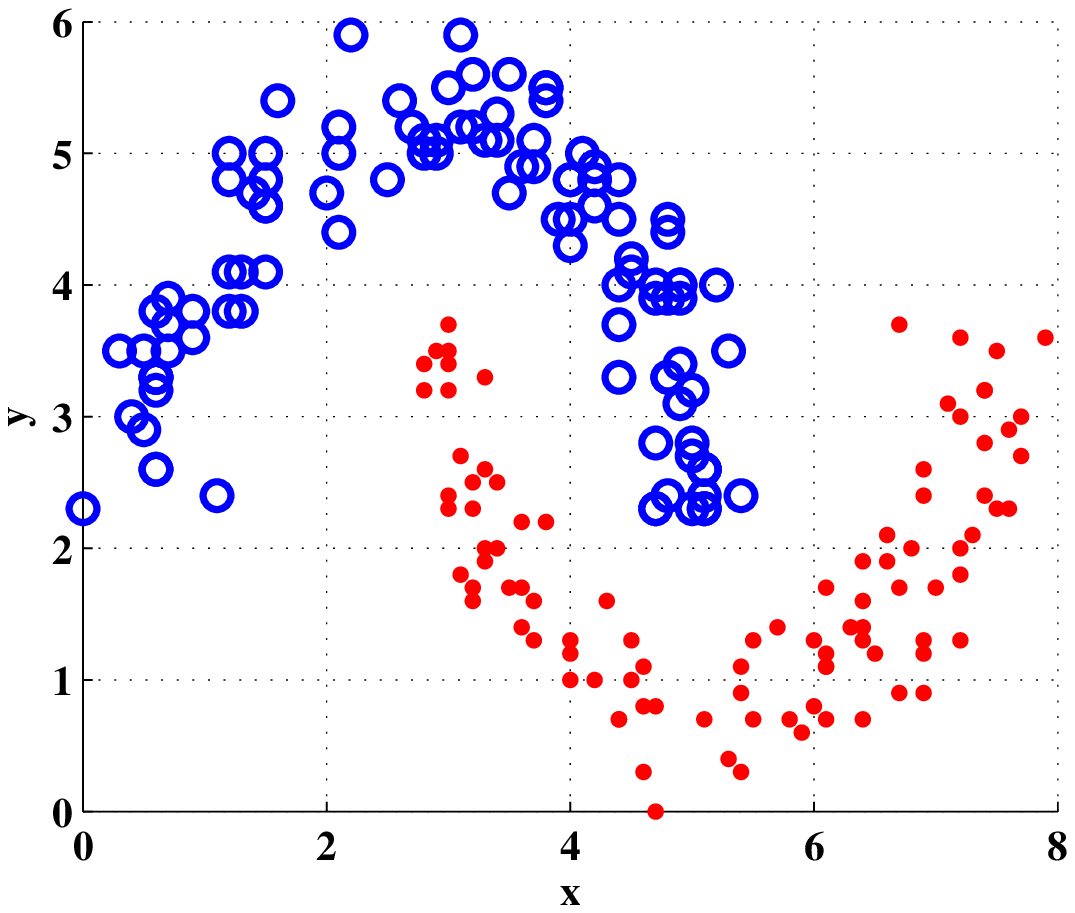}}
\subfigure[Data set \#3]{
\label{fig:5-15} 
\includegraphics[width=3.6cm,height=3.4cm,origin=c]{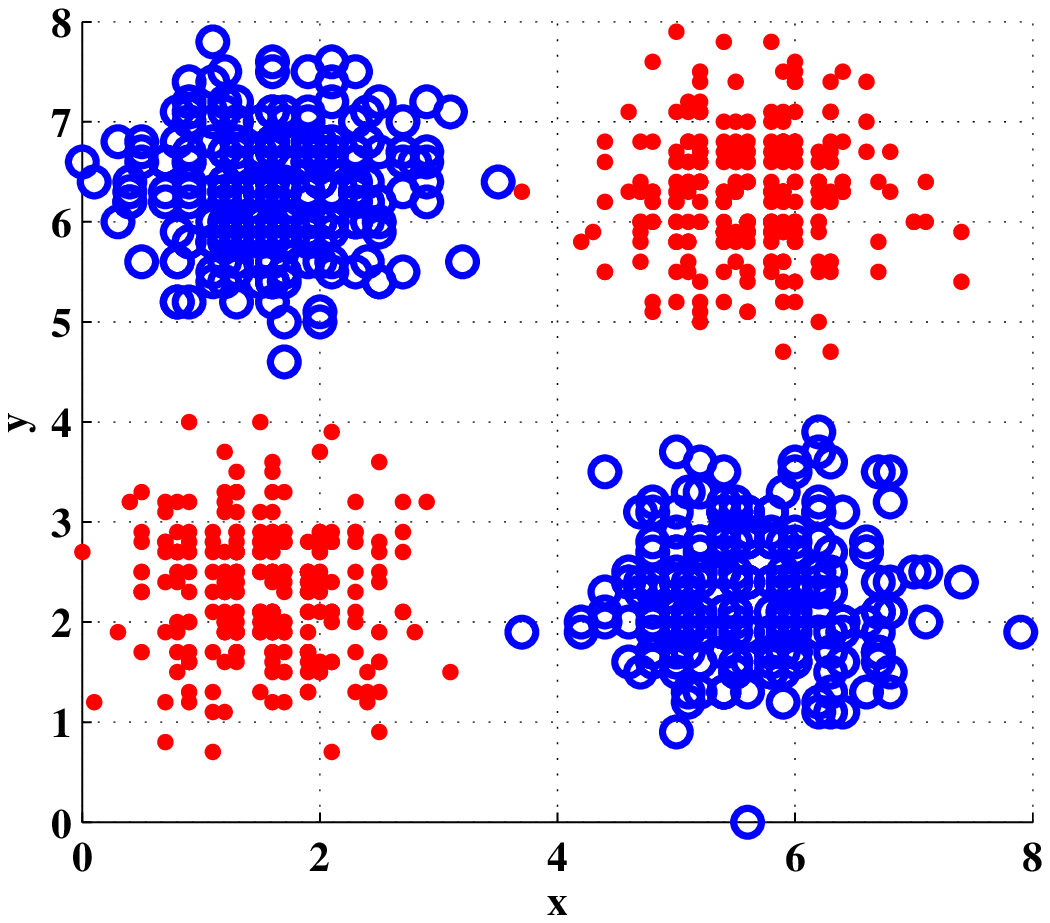}}
\subfigure[Data set \#4]{
\label{fig:5-16} 
\includegraphics[width=3.6cm,height=3.4cm,origin=c]{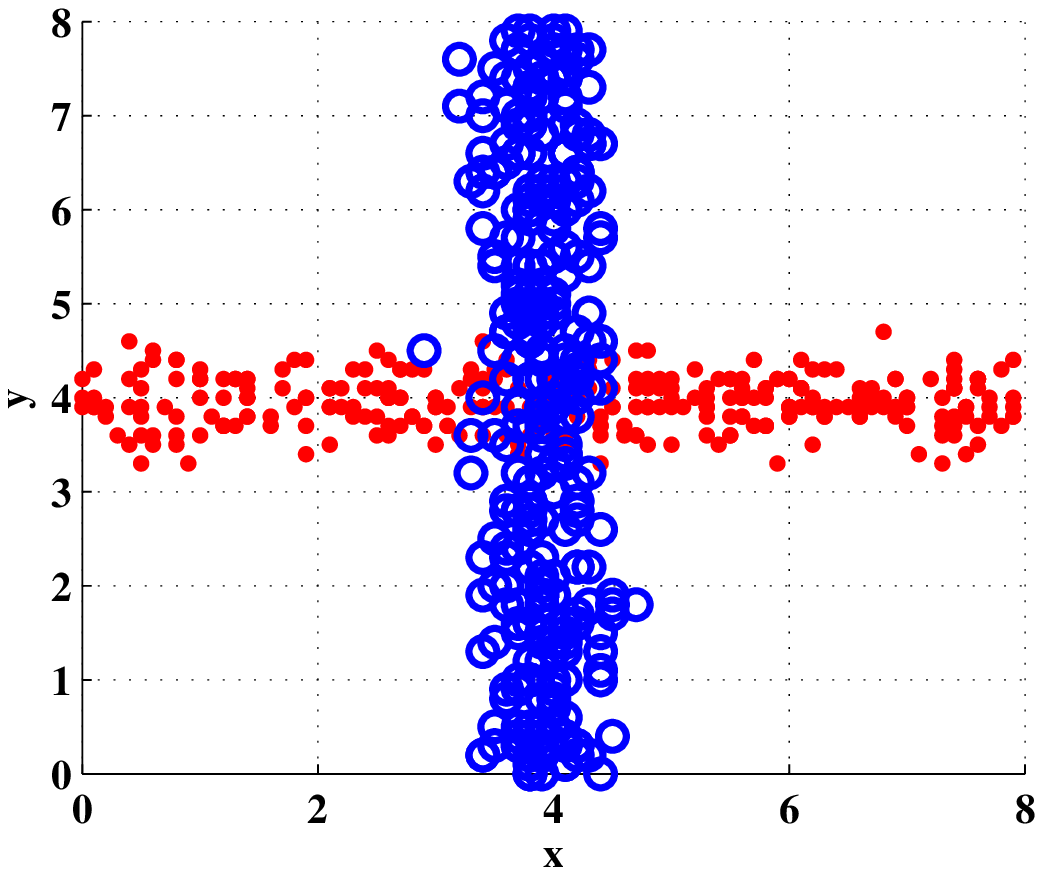}}
\caption{Data sets used to demonstrate the ability of the proposed
neuro-fuzzy system for data classification.}
\label{fig5-13and14and15and16} 
\end{figure*}

\begin{figure*}[!t]
\centering \subfigure[Data set \#1]{
\label{fig:5-17} 
\includegraphics[width=3.6cm,height=3.4cm]{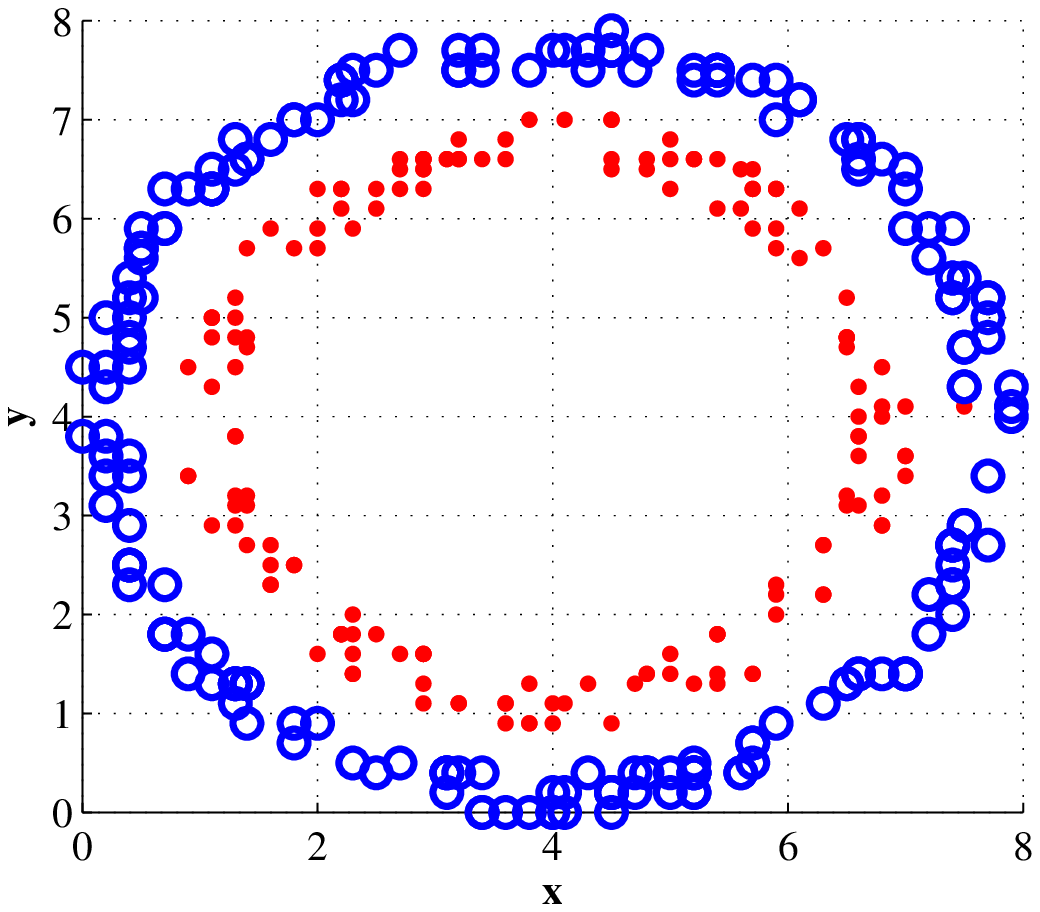}}
\subfigure[Data set \#2]{
\label{fig:5-18} 
\includegraphics[width=3.6cm,height=3.4cm,origin=c]{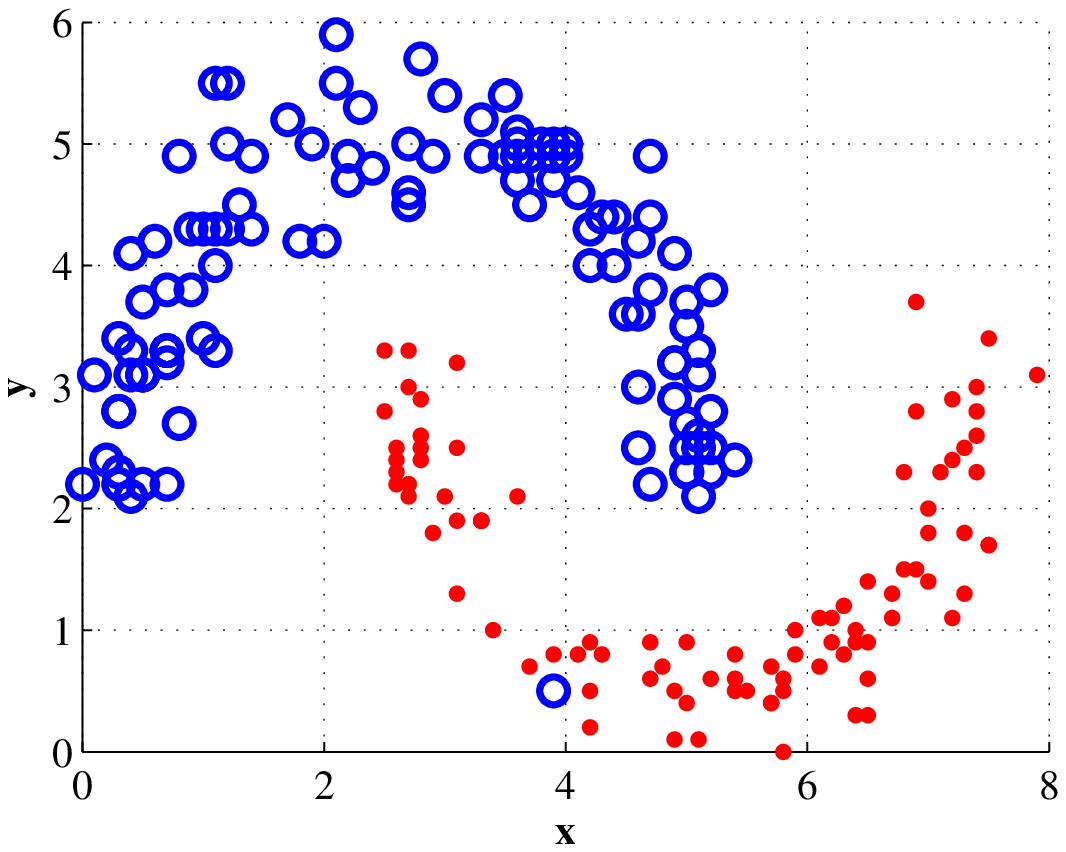}}
\subfigure[Data set \#3]{
\label{fig:5-19} 
\includegraphics[width=3.6cm,height=3.4cm,origin=c]{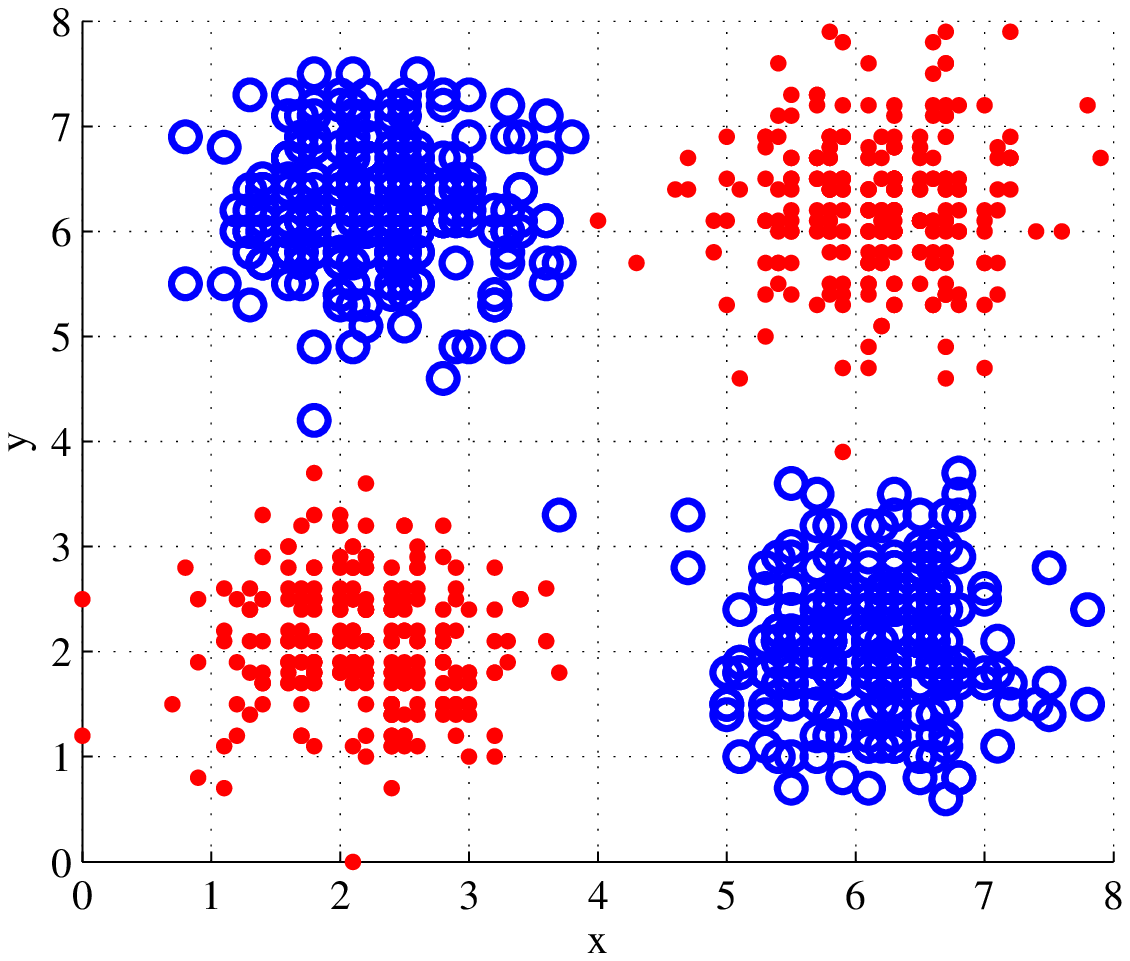}}
\subfigure[Data set \#4]{
\label{fig:5-20} 
\includegraphics[width=3.6cm,height=3.4cm,origin=c]{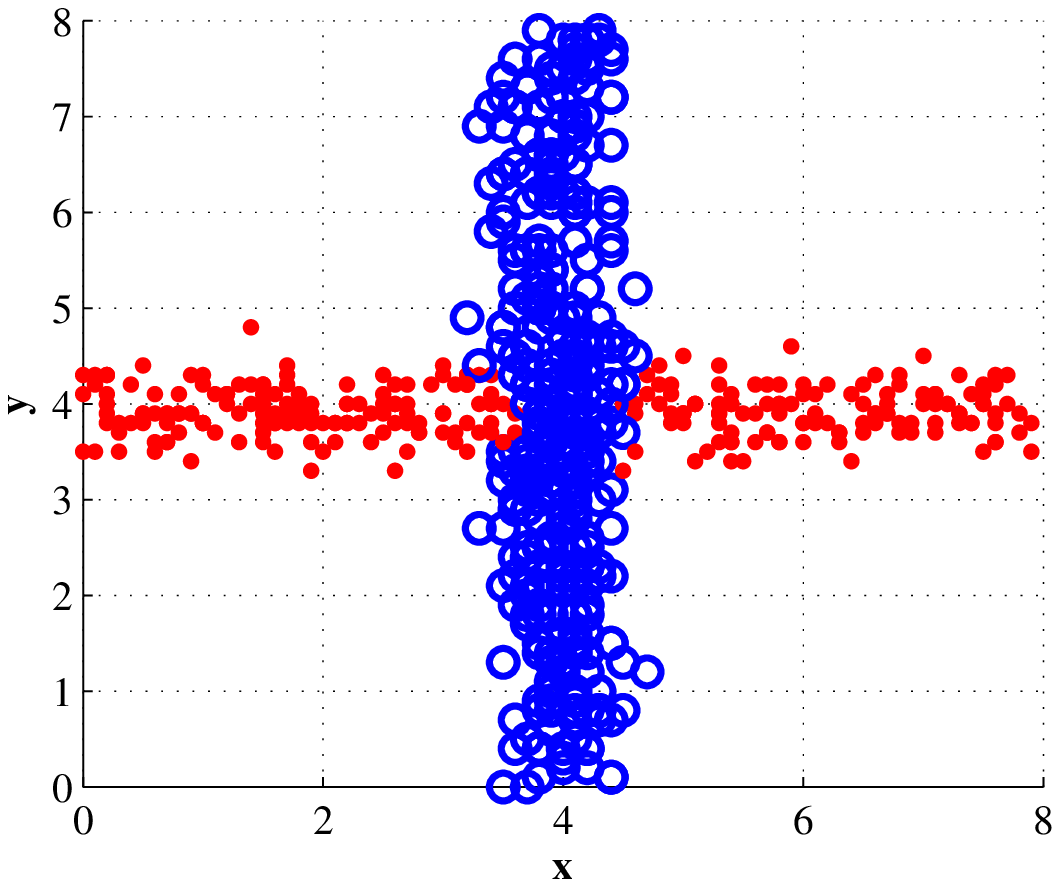}}
\caption{This figure shows an example of how the new
data points have been classified by the trained system.}
\label{fig5-17and18and19and20} 
\end{figure*}

In the following we discuss on the ability of the proposed
neuro-fuzzy system for data classification. The reason for the
importance of this problem is that it shows the ability of 
system for making suitable decisions in facing with new
environmental inputs. For this purpose consider four different
data sets (each consists of two different objects) as shown in
Fig. \ref{fig5-13and14and15and16}. In each case first we train the system shown in Fig. \ref{fig3}
with only two output terminals (one per each class) using supervisory method. Then we present the new data points to
system and let it classify them. Table \ref{tableclass} shows the details of the parameters used in simulations and
the results obtained at each case. Figure \ref{fig5-17and18and19and20} shows an example of how the new
data points have been classified by the trained system. It is also
concluded from Table \ref{tableclass} that in all cases only a small fraction of
the whole training data is sufficient to construct a neuro-fuzzy
computing system with the ability of data classification with a
high precision.

\begin{table*}[ht]
\small
\caption{Classification rate (\%) for each test set} 
\centering 
\begin{tabular}{p{1.4cm}| p{1.9cm} p{1.9cm} p{1.5cm} p{1.6cm} p{1.9cm}} 
\hline\hline \hspace{0.2cm}
Data set & \# of neurons for variable $x$ & \# of neurons for variable $y$ & \# of training data & \# constructed minterms& classification rate (\%) \\
\hline
 1 & 100 & 100 & 335 & 52 & 99.8 \\
 2 & 90 & 90 & 200 & 45 & 99.36\\
 3 & 98 & 98 & 1000 & 107 & 99.64\\
 4 & 94 & 94 & 600 & 40 & 95.83\\
\hline
\end{tabular}
\label{tableclass} 
\end{table*}


Finally, we can study the performance of the proposed neuro-fuzzy
computing system when it is trained with the noisy data or when a
certain percent of the memristors at cross-points are distorted
randomly. First consider the problem of modeling $g_1$,...,$g_5$
(as defined in (\ref{g1})-(\ref{g5})) when the system is trained
with the noisy data. For this purpose, we add a Gaussian noise
with zero average and variance of 0.01 to each of the 225
input-output training data, and then we train the system shown in
Fig. \ref{fig3} using these noisy data (recall that the original data are
randomly selected (with uniform distribution) from the domain of
definition of the corresponding functions). Table \ref{tablenoise} shows the
values obtained for FVU under the above-mentioned conditions in a
typical simulation. These results show the good robustness of the
proposed system in dealing with noisy data. 
Repeating the above simulation assuming that the training data is
not noisy but 20 percent of the cross-points are randomly
distorted beforehand and cannot be used to save data, we arrive at
the results presented in Table \ref{tabledefective} (a randomly chosen value is
assigned to each distorted memristor during simulation). These
results clearly show that the proposed system can fairly tolerate
such a distortion. In fact, in this case the number of fuzzy
min-terms is automatically increased to improve the accuracy of
system.

\begin{table}[ht]
\small
\caption{Noise tolerance of models using FVU criterion} 
\centering 
\begin{tabular}{c| c c c c c} 
\hline\hline \hspace{0.2cm}
Function & $g_1$ & $g_2$ & $g_3$ & $g_4$ & $g_5$ \\[1ex]
\hline\hspace{0.5cm}
FVU & 0.281 & 0.394 & 0.727 & 0.586 & 0.61
\end{tabular}
\label{tablenoise} 
\end{table}

\begin{table}[ht]
\small
\caption{Fault tolerance of the proposed structure in modeling application using FVU criterion} 
\centering 
\begin{tabular}{c| c c c c c} 
\hline\hline \hspace{0.2cm}
Function & $g_1$ & $g_2$ & $g_3$ & $g_4$ & $g_5$ \\[1ex]
\hline\hspace{0.5cm}
FVU & 0.212 & 0.096 & 0.357 & 0.144 & 0.228\\[1ex]
\# of constructed min-terms & 96 & 186 & 156 & 152 & 139\\
\hline
\end{tabular}
\label{tabledefective} 
\end{table}

\section{Conclusion}\label{conclu}
One novelty of this paper is in the new explanation presented for
the relation between logical circuits and ANNs, logical circuits
and fuzzy logic, and ANNs and fuzzy inference systems. This
explanation led to the notion of \emph{fuzzy min-terms} and a
special two-layer ANN with the capacity of working with fuzzy
input-output data. This neuro-fuzzy computing system has at least
four main advantages compared to many other classical designs.
First, it can effectively be realized on the nano-scale
memristor-crossbar structure. Second, the hardware of system can
be trained simply by applying the Hebbian learning method and
without the need to any optimization. Third, the proposed
structure can effectively work with huge number of input-output
training data (of fuzzy type) without facing with problems like
overtraining. Finally, it has a strong biological support, which
makes it a powerful structure to emulate the function of human
brain.


%





\ifCLASSOPTIONcaptionsoff
  \newpage
\fi

\end{document}